\documentclass[sigconf]{acmart}
\usepackage{hyperref}
\usepackage{url}
\usepackage{multirow}
\usepackage{array}
\usepackage{booktabs} 
\usepackage{graphicx}
\usepackage{bm}
\usepackage{enumitem}
\usepackage{adjustbox}
\usepackage{footmisc}
\newtheorem*{remark}{Remark}
\setlist[itemize]{leftmargin=*}

\AtBeginDocument{%
  \providecommand\BibTeX{{%
    \normalfont B\kern-0.5em{\scshape i\kern-0.25em b}\kern-0.8em\TeX}}}

\copyrightyear{2024}
\acmYear{2024}
\setcopyright{acmlicensed}
\acmConference[KDD '24]{Proceedings of the 30th ACM SIGKDD Conference on Knowledge Discovery and Data Mining}{August 25--29, 2024}{Barcelona, Spain}
\acmBooktitle{Proceedings of the 30th ACM SIGKDD Conference on Knowledge Discovery and Data Mining (KDD '24), August 25--29, 2024, Barcelona, Spain}
\acmDOI{10.1145/3637528.3671957}
\acmISBN{979-8-4007-0490-1/24/08}
\begin{document}

\title{Relaxing Continuous Constraints of Equivariant \\Graph Neural Networks for Physical Dynamics Learning}


\author{Zinan Zheng}
\authornote{Equal contribution. Listing order is random. Yang Liu designed and implemented the first DEGNN.  Zinan designed, implemented and evaluated countless model variants.}
\affiliation{%
  \institution{Hong Kong University of Science and\\
 Technology (Guangzhou)}
 \country{}
}
\email{zzheng078@connect.hkust-gz.edu.cn}

\author{Yang Liu}
\authornotemark[1]
\affiliation{%
  \institution{Hong Kong University of Science and
 Technology}
 \institution{Hong Kong University of Science and\\
 Technology (Guangzhou)}
  \country{}
}
\email{yliukj@connect.ust.hk}

\author{Jia Li}
\authornote{Corresponding author.}
\affiliation{%
  \institution{Hong Kong University of Science and\\
 Technology (Guangzhou)}
 \country{}
}
\email{jialee@ust.hk}

\author{Jianhua Yao}
\author{Yu Rong}
\authornotemark[2]
\affiliation{%
  \institution{Tencent AI Lab}
  \country{}
}
\email{jianhua.yao@gmail.com}
\email{yu.rong@hotmail.com}

\renewcommand{\shortauthors}{Zinan Zheng, Yang Liu, Jia Li, Jianhua Yao, and Yu Rong}

\begin{abstract}
Incorporating Euclidean symmetries (e.g. rotation equivariance) as inductive biases into graph neural networks has improved their generalization ability and data efficiency in unbounded physical dynamics modeling. However, in various scientific and engineering applications, the symmetries of dynamics are frequently discrete due to the boundary conditions. Thus, existing GNNs either overlook necessary symmetry, resulting in suboptimal representation ability, or impose excessive equivariance, which fails to generalize to unobserved symmetric dynamics.
In this work, we propose a general \textbf{D}iscrete \textbf{E}quivariant \textbf{G}raph \textbf{N}eural \textbf{N}etwork (DEGNN) that guarantees equivariance to a given discrete point group. Specifically, we show that such discrete equivariant message passing could be constructed by transforming geometric features into permutation-invariant embeddings. Through relaxing continuous equivariant constraints, DEGNN can employ more geometric feature combinations to approximate unobserved physical object interaction functions. 
Two implementation approaches of DEGNN are proposed based on ranking or pooling permutation-invariant functions.
We apply DEGNN to various physical dynamics, ranging from particle, molecular, crowd to vehicle dynamics. In twenty scenarios, DEGNN significantly outperforms existing state-of-the-art approaches.  
Moreover, we show that DEGNN is data efficient, learning with less data, and can generalize across scenarios such as unobserved orientation.
\end{abstract}

\ccsdesc[500]{Computing methodologies~Machine learning}


\keywords{Equivariant graph neural network, physical dynamics}

\maketitle

\section{Introduction}

\begin{figure}[t]
    \centering
    \includegraphics[width=0.4\textwidth]{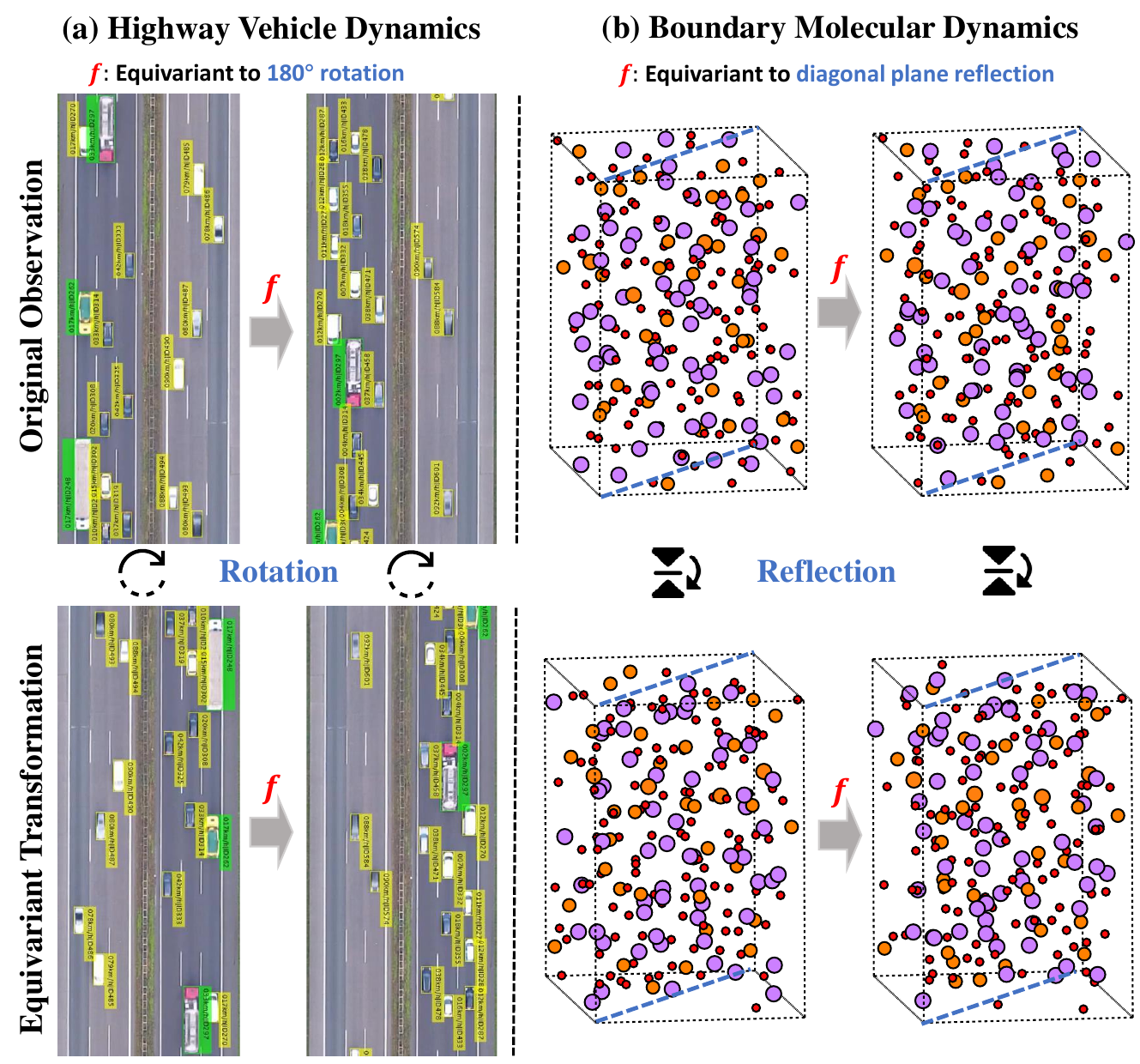}
    \vspace{-3ex}
    \caption{Illustration of discrete equivariance. (a) Rotation equivariance of vehicle trajectories. They have similar patterns when they drive in lanes of opposite directions; (b) Reflection equivariance of molecular dynamics. The effects of boundaries and molecules are symmetric when we reflect the entire system.}
    \vspace{-3ex}
    \label{fig:intro}
\end{figure}

Accurate physical dynamic modeling is a fundamental task in numerous applications. For example, understanding crowd (i.s., a group of pedestrians) dynamics is essential for public safety, urban planning, and architecture design~\cite{johansson2008crowd, DBLP:journals/cvgip/YangLGPH20,DBLP:journals/tits/KothariKA22}. In scientific domains, simulating molecular dynamics is pivotal in material science~\cite{DBLP:conf/nips/KabaR22,li2022graph}, drug discovery~\cite{durrant2011molecular}, and protein folding~\cite{DBLP:conf/nips/Han0XR22}. These systems have complex latent behavior patterns (e.g., agents' intentions or physical laws), which are generally difficult to observe. Consequently, the object interactions are either intractable (e.g., social interactions among pedestrians)~\cite{DBLP:conf/cvpr/AlahiGRRLS16,liu2023human} or with high computation complexity (e.g., interactions between atoms or proteins)~\cite{unke2021machine,yang2021hierarchical,feng2024deep,gao2023hierarchical,huang2023generalizing,han2024survey}. 
In particular, the core aspect of their behavior forecasting is representing and reasoning the interactions among system objects. To achieve this, multiple Graph Neural Networks (GNNs)~\cite{DBLP:conf/www/ShiYL23,DBLP:conf/icml/Sanchez-Gonzalez20,satorras2021n,DBLP:conf/iclr/0001HRX0H22,ma2022cross,liu2023improving,li2023survey,zhao2023effective,zhao2024weakly,zhao2024all,han2024survey,yang2021hierarchical} have been proposed for learning interactions of various physical systems such as crowds, particles, and molecules. They represent system objects as nodes, physical relations as edges, and their interactions as the message passing thereon.

Basically, physical dynamics demonstrates particular symmetries. Figure~\ref{fig:intro} displays two examples. Vehicles exhibit analogous behavior patterns on two opposing directions highways, and the trajectories of molecules remain equivariant under reflection. 
Regardless of such inductive bias, current methods such as GNS~\cite{DBLP:conf/icml/Sanchez-Gonzalez20} and Transformer~\cite{giuliari2021transformer} fail to generalize to unobserved directions. Considering highway vehicle dynamics, if training data are in a downward direction, GNS can predict well during testing when the vehicles are in the same direction but perform poorly if the vehicles are from the bottom up (See Section~\ref{sec:rq2}). This observation implies that conventional GNNs are insufficient for capturing the true dynamics and tend to overfit the observed trajectories.

Therefore, geometrically equivariant graph neural networks~\cite{DBLP:journals/corr/abs-2202-07230} have been proposed to leverage symmetry as an inductive bias to efficiently model physical dynamics. They force their outputs to be strictly equivariant under a given group, e.g., SE(3) (rotation and translation)~\cite{thomas2018tensor, fuchs2020se} and E(3)~\cite{DBLP:conf/iclr/BrandstetterHPB22, batzner20223} (rotation, translation, and reflection). 
However, strict continuous symmetry is rarely observed in physical dynamics. The widely existing boundary, such as sidewalks, highways, and boundary conditions in molecular simulation, generally breaks such symmetry into discrete elements. Learning such dynamics is highly challenging due to (1) \textbf{Discrete symmetry}. GNNs equivariant to continuous groups, like EGNN~\cite{satorras2021n} and Eqmotion~\cite{xu2023eqmotion}, are so strong and overly restrictive, limiting their flexibility and hindering their potential applications. Relaxing the continuous assumption and building GNNs that are equivariant to a \textit{discrete group} is necessary; (2) \textbf{Intricate interaction}, which is divergent and dense. Specifically, the underlying interactions of different physical systems are diverse. In addition, we consider high-density scenarios where interactions occur with great frequency. Therefore, the model needs to be flexible enough to approximate these interactions.

In this paper, we propose a novel \textbf{D}iscrete \textbf{E}quivariant \textbf{G}raph \textbf{N}eural \textbf{N}etwork framework, called DEGNN, to forecast various types of physical dynamics.
It first augments the input geometric features with transformations of the given group and then employs a permutation-invariant function, followed by radial direction multiplication, to obtain the message embedding. 
We show that such a general framework is equivariant to a given point group. 
In addition, since it relaxes the constraints on geometric features, it could utilize more expressive feature combinations to approximate various types of object interactions.
Under this framework, we propose two realizations, ranking and pooling methods. 
Extensive experiments on four physical systems, including macro-level vehicles and crowds, and micro-level particle and molecular dynamics, demonstrate that DEGNN has a better generalization ability over state-of-the-art models and is beneficial for learning the discrete symmetry of dynamics.

Our main contributions are summarized as follows:
\begin{itemize}
    \item We highlight the importance of learning discrete equivariance in physical dynamic modeling, which is challenging and vital in scientific and engineering applications. To the best of our knowledge, it is the first time discrete symmetry learning is considered in such domains.
    \item We propose a novel and general DEGNN framework that is proven to be equivariant to discrete groups. 
    \item We propose two realizations of the DEGNN framework, via permutation-invariant ranking and pooling functions.
    \item We conduct extensive experiments to evaluate the performance of DEGNN on \textbf{twenty} scenarios of four types of physical systems. Experimental results show that the proposed model achieves significantly better performance than the state-of-the-art models. Additional results including ablation studies, generalization experiments, and sensitivity analysis further demonstrate the generalization ability of DEGNN. The code is available at the link: \url{https://github.com/compasszzn/DEGNN}.
\end{itemize}

\section{Preliminary}
\subsection{Problem Definition}
In this work, we study various physical dynamics that consist of complex interacting objects (e.g., atoms, pedestrians, and vehicles) in static and bounded environments, and their trajectories are governed by complex physical rules or social interactions (e.g., vehicles tend to avoid collision). Therefore, our goal is to learn interactions of $N$ objects and forecast their positions after a fixed time interval. At time $t$, each object $i$ is represented by: 
\begin{itemize}
    \item Geometric features including the position vector $\bm{q}_{i}^{(t)} \in\mathbb{R}^{2}$ and the velocity vector $\bm{\dot{q}}_{i}^{(t)}\in\mathbb{R}^{2}$.
    \item Non-geometric features such as the atom number or vehicle type, denoted by $\bm{u}_{i}$.
    \item Spatial connection where an edge $e_{ij}$ is constructed via distance cutoff or physical relations. Generally, edges are associated with attributes such as object distance, denoted by $a_{ij}$.
\end{itemize} 
For simplicity, we denote $(\bm{q}^{(t)}, \bm{\dot{q}}^{(t)})$ and $(\bm{u}, \bm{e} = \{ e_{ij} \}, \bm{a} = \{ a_{ij} \} )$ as dynamic and static state information of the entire graph correspondingly. Formally, the dynamic forecasting problem is defined as follows:
\begin{definition}
(Dynamic Forecasting) Given the initial system states $(\bm{q}^{(t)}, \bm{\dot{q}}^{(t)})$ at time $t$ and static states $(\bm{u}, \bm{e}, \bm{a})$, the objective is to predict the subsequent position $\bm{q}^{(t+\Delta t)}$, where $\Delta t$ is the target time interval.
\end{definition}

\subsection{Equivariance and Invariance}
\subsubsection{\textbf{Group}}
Equivariance and invariance play a pivotal role in physical dynamic modeling.
Formally, it is defined on a specific group~\cite{suzuki1986group}: 
\begin{definition}
(Group) A group is a set of operations that satisfy: closure, associativity, the existence of an identity element, and the existence of inverse elements for each element in the set.
\end{definition}
\noindent
For instance, a widely studied group in geometric graph learning is Euclidean group $E(n)$, which includes transformations of translation, rotation, and reflection. 

\subsubsection{\textbf{Equivariance and invariance}}
Given a group $G$, the definition of $G$-Equivariance and $G$-Invariance is:
\begin{definition}
($G$-Equivariance and $G$-Invariance) A function $f$ is equivariant to group $\mathcal{G}$, if for any transformation $g \in \mathcal{G}$, $f(g\circ x) = g \circ f(x), \quad x\in \bm{X}$.

Similarly, $f$ is invariant to group $\mathcal{G}$, if for any transformation $g \in \mathcal{G}$, $f(g\circ x) = f(x), \quad x\in \bm{X}$.

\end{definition}
\noindent
The group transformation $\circ$ is instantiated as $g\circ x := \bm{O}x + \bm{t}$ where $\bm{O}\in O(n):=\{\bm{O}\in\mathbb{R}^{n\times n}|\bm{O}^{T}\bm{O}=\bm{I}\}$ is orthogonal transformation (rotation and reflection), and $\bm{t}\in\mathbb{R}^{n}$ is a translation vector.

\subsubsection{\textbf{Permutation-invariance}}
Another important inductive bias is permutation-invariance. That is, the function output remains unchanged regardless of the order in which its inputs are presented. Formally, given the input is a set $\{x_1. x_2, \cdots, x_d\}$ where $d$ is the size, its definition is:
\begin{definition}
(Permutation-invariance) A function acting on sets is permutation-invariant, if for any permutation $\pi$,
\begin{equation}
\begin{aligned}
    f(x_1, x_2,\cdots, x_d) = f(x_{\pi(1)}, x_{\pi(2)},\cdots, x_{\pi(d)}).
\end{aligned}
\end{equation}
\end{definition}

\subsection{Equivariant Message Passing}
Since the composition of equivariant functions is again equivariant, equivariant GNNs~\cite{satorras2021n,DBLP:conf/iclr/0001HRX0H22} are built via stacking multiple equivariant message-passing layers. Thus, their output will change in the same way as the input changes. In general, their $l$-th layer computes:
\begin{equation}
\begin{aligned}
    \bm{m}_{ij}^{(l)}  = \mu(\bm{q}_{i}^{(l)}, \bm{q}_{j}^{(l)}, \bm{\dot{q}}_{i}^{(l)}, \bm{\dot{q}}_{j}^{(l)}, \bm{h}_{i}^{(l)}, \bm{h}_{j}^{(l)}, a_{ij}), \\
    \bm{q}_{i}^{(l+1)}, \bm{\dot{q}}_{i}^{(l+1)}, \bm{h}_{i}^{(l+1)}  = \nu(\bm{q}_{i}^{(l)}, \bm{\dot{q}}_{i}^{(l)}, \bm{h}_{i}^{(l)}, \sum_{j\in \mathcal{N}_i}\bm{m_{ij}}^{(l)}),
\end{aligned}
\end{equation}
where $\bm{h}_{i}^{(l)}$ is the $l$-th layer embedding of node $i$ and $\bm{m}_{ij}^{(l)}$ denotes the $l$-th layer message embedding between node $i$ and node $j$. $\mathcal{N}_i$ collects the neighbors of node $i$. $\mu$ and $\nu$ are the equivariant message embedding function and node state updating function, respectively. 

\subsubsection{\textbf{EGNN framework}}
It is non-trivial to design the above equivariant message embedding functions. A feasible solution is based on the EGNN framework~\cite{satorras2021n}, which relies on the inner product to transform geometric vectors into invariant features, followed by an MLP and radial directions multiplication. They have the following form:
\vspace{-1ex}
\begin{equation}
\begin{aligned}
    \mu(\bm{x}, \bm{h}) = \bm{x}\sigma(\bm{x}^{T}\bm{x}, \bm{h}),
\end{aligned}\label{eq:inv}
\end{equation}
where $\bm{x},\bm{h}$ denotes abbreviated geometric and non-geometric terms and $\bm{\sigma}$ is an MLP.
For example, EGNN employs the relative squared distance as the invariant feature:
\vspace{-1ex}
\begin{equation}
\begin{aligned}
    \mu_{\text{egnn}}(\bm{q}_{i}, \bm{q}_{j}, \bm{h}) = (\bm{q}_{i} - \bm{q}_{j})\sigma(||\bm{q}_{i}-\bm{q}_{j}||^2, \bm{h}),
\end{aligned}
\end{equation}
due to that for any orthogonal transformation $\bm{O}$ ($\bm{O}^{T}\bm{O}=\bm{I}$) and translation vector $\bm{t}$ in $E(n)$:
\begin{equation}
\begin{aligned}
    ||(\bm{O}\bm{q}_{i} + \bm{t}) - (\bm{O}\bm{q}_{j} + \bm{t})||^2=||\bm{O}(\bm{q}_{i} - \bm{q}_{j})||^2 = ||\bm{q}_{i} - \bm{q}_{j}||^2.
\end{aligned}
\end{equation}
Then by multiply the MLP output with $\bm{q}_{i} - \bm{q}_{j}$, $\mu_{\text{egnn}}$ is rotation and reflection equivariant.
Since involving $E(n)$-invariant features will not change its equivariant property, other feasible features are the inner product of velocity~\cite{DBLP:conf/nips/KabaR22} or expanding the relative distance with radial basis functions~\cite{DBLP:conf/iclr/WangC23}.

Given the equivariant message embedding, the node states are updated through a physics-inspired integration:
\begin{equation}
\begin{aligned}
    \bm{\dot{q}}_{i}^{(l+1)}  &= \psi(\bm{h}_{i}^{(l)})\bm{\dot{q}}_{i}^{(l)} + \sum_{j\in \mathcal{N}_i}\bm{m_{ij}}^{(l)}, \\
    \bm{q}_{i}^{(l+1)} = \bm{q}_{i}^{(l)} + &\bm{\dot{q}}_{i}^{(l+1)},\quad\bm{h}_{i}^{(l+1)} = \sigma_h(\bm{q}_{i}^{(l)}, \sum_{j\in \mathcal{N}_i}\bm{m_{ij}}^{(l)})
\end{aligned}~\label{eq:update}
\end{equation}
where $\psi(\bm{h}_{i}^{(l)})\in\mathbb{R}$ is a scalar that controls the magnitude of velocity. The non-geometric features could be updated via $\bm{h}_{i}^{(l+1)} = \sigma_h(\bm{q}_{i}^{(l)}, \sum_{j\in \mathcal{N}_i}\bm{m_{ij}}^{(l)})$.
However, such frameworks follow the constraint of continuous groups and are not straightforward to be relaxed to a discrete group.
\subsubsection{\textbf{Model training}}
The final prediction is obtained by applying several iterations of equivariant message passing. Then the model parameters can be optimized by minimizing the discrepancy between exact and approximated positions:
\begin{equation}
\begin{aligned}
    \mathcal{L}_{\text{train}}= \sum_{s\in\mathcal{D}_{\text{train}}}||\bm{\hat{q}}_{s}^{(t + \Delta t)}-\bm{q}_{s}^{(t + \Delta t)}||_{2},
\end{aligned}
\label{eq:loss}
\end{equation}
where $\mathcal{D}_{\text{train}}$ denotes the training set. $\bm{\hat{q}}_{s}^{(t + \Delta t)}, \bm{q}_{s}^{(t + \Delta t)}$ are the model prediction and actual position of sample $s$.

\section{Method}
\subsection{Framework}
The core of the EGNN framework lies in the invariant inner product. Thus, our goal is to replace it via an invariant function to point group and consequently relax the $E(n)$ equivariant message passing. In this section, we first introduce the point group and then elaborate on our equivariant message-passing layer.

\subsubsection{\textbf{Point group}}
The symmetry of dynamics is highly affected by their boundary conditions, which only have finite discrete elements in many circumstances. Mathematically, it can be described by a point group $P$~\cite{bradley2010mathematical}, the set of isometries that maps the boundary structure to itself. As isometries, point groups are subgroups of the orthogonal group $O(n)$, which will be rotations and reflections. Different symmetric boundaries correspond to different point groups. For example, the symmetry of a square is described by the $D4$ group, which consists of identity, 90°/180°/270° rotation, horizontal/vertical reflection, and reflection of two diagonal lines, and the point group of a non-rectangular parallelogram is $D_1$, which only contains the identity and 180° rotation. 

\subsubsection{\textbf{$P$-equivariant message embeddings}}
It is challenging to derive a straightforward $P$-equivariant function. Nevertheless, note that the inner product in Eq.~\ref{eq:inv} is essentially a function $\phi:\{\bm{O}\bm{x}+\bm{t}\}_{\bm{O},\bm{t}\in E(n)}\rightarrow \mathbb{R}^{n}$ that maps all transformed geometric vectors $\bm{O}\bm{x}+\bm{t}$ in $E(n)$ to the same value $\bm{x}^{T}\bm{x}$. Inspired by this insight, we relax equivariance to a point group $P$ via only reducing the function input to transformed geometric vectors in $\{\bm{O}\bm{x}\}_{\bm{O}\in P}$. Specifically, we first enhance the input using the point group to generate a set of transformed geometric vectors. Then, the function $\phi$ converts this set of vectors into the same embedding. 
Consequently, the $P$-equivariant function $\mu_P$ is defined as follows:
\begin{equation}
\begin{aligned}
    \mu_P(\bm{x}, \bm{h}) = (\bm{q}_i - \bm{q}_j)\phi(\{\bm{O}\bm{x}\}_{\bm{O}\in P}, \bm{h}).
\end{aligned}\label{eq:point}
\end{equation}
Here $\bm{x}$ is \textbf{symmetry-breaking} features. For example, they can be $(\bm{q}_{i}, \bm{q}_{j}, \bm{\dot{q}}_{i}, \bm{\dot{q}}_{j})$ or $(\bm{q}_{i}, \bm{q}_{j}, \bm{\dot{q}}_{i}, \bm{\dot{q}}_{j}, \bm{q}_{i} - \bm{q}_{j}, ||\bm{q}_{i} - \bm{q}_{j}||^2)$. Such construction leads to $P$-equivariance, otherwise employing a set of $E(n)$-invariant features (e.g., $||\bm{x}_i - \bm{x}_j||^{2}$ or $||\bm{v}_i||^2$) will enhance the symmetry to $E(n)$-equivariance.

\begin{remark}
    The feature $\{\bm{O}\bm{x}\}_{\bm{O}\in P}$ is invariant to the point group $P$.
\end{remark}
Since the set is permutation-invariant and the group is closure (the result of two elements within a group is still an element of the group), for any transformation $\bm{O}\in P$, the input $\bm{x}$ and $\bm{Ox}$ result in the same $\{\bm{O}\bm{x}\}_{\bm{O}\in P}$. That is, let $\bm{O}_1, \bm{O}_2\in P$ be transformations in point group $P$, we have $\bm{O}_1\bm{O}_2\bm{x}\in \{\bm{O}\bm{x}\}_{\bm{O}\in P}$.
Thus, we immediately have the following theory:
\begin{theorem}\label{thm:equ}
For arbitrary rotation or reflection matrix $\bm{O}\in P$, if $\phi$ is a permutation-invariant function, the function $\mu_P$ satisfies the $P$-equivariance:
\begin{equation}
\begin{aligned}
    \mu_P(\bm{O}\bm{x}, \bm{h}) = \bm{O}\mu_P(\bm{x}, \bm{h}).
\end{aligned}
\end{equation}
\end{theorem}
\noindent
The proof is provided in Appendix~\ref{proof:thm} where we show that $\phi$ is invariant to $P$ and by multiplying it with radial directions $\bm{q}_i - \bm{q}_j$, we achieve $P$-equivariant message embedding. 

Finally, following the same equivariant node updating functions (i.e., Eq.~\ref{eq:update}) and training strategy, we obtain a message-passing layer equivariant to a given point group $P$.

\begin{figure}[t]
    \centering
    \includegraphics[width=0.45\textwidth]{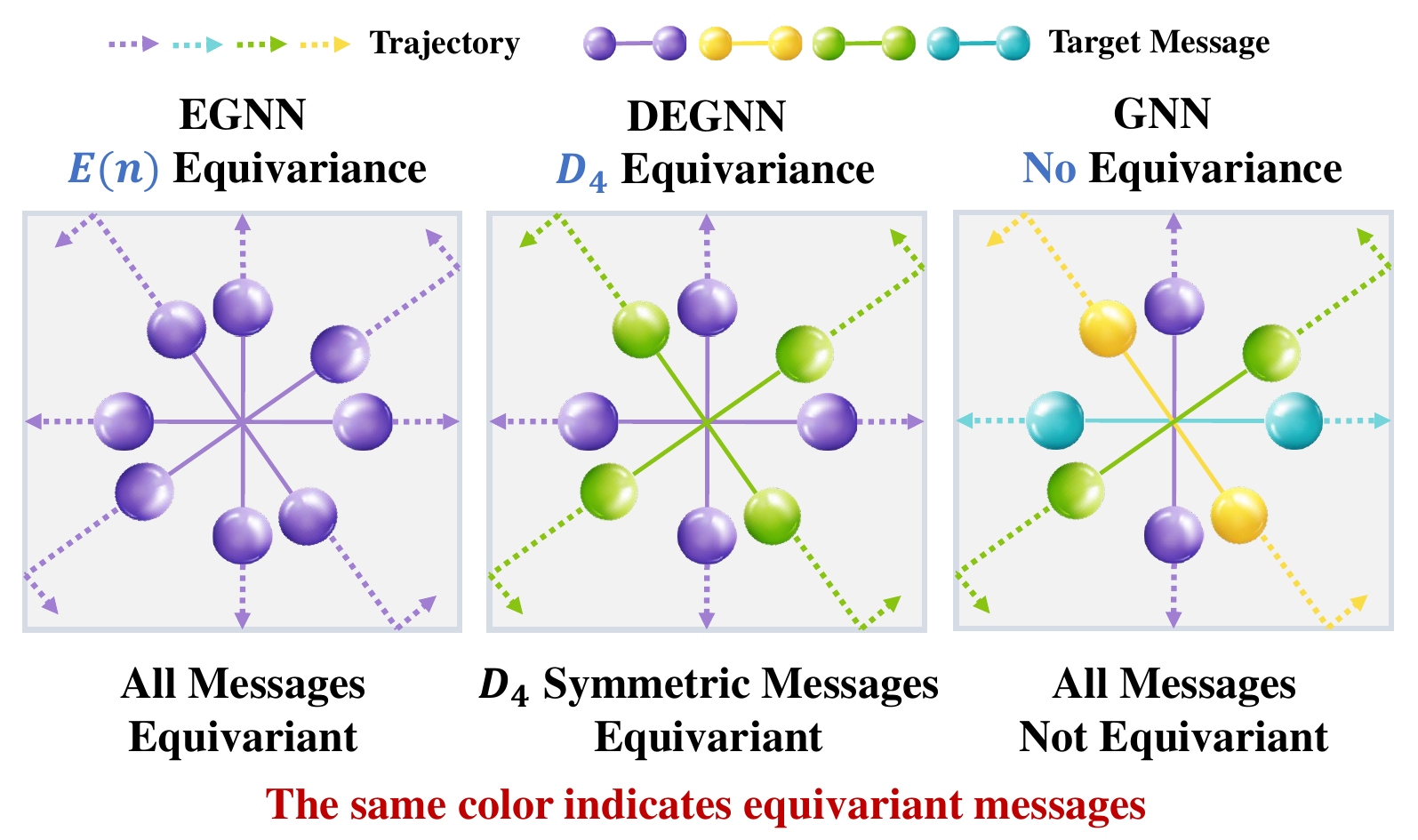}
    \vspace{-1ex}
    \caption{Examples of N-body system trajectories. The square is the boundary where its symmetry is described by the $D_4$ group, a specific point group. EGNN and GNN fail to distinguish the interactions of the objects that result in different dynamics, while DEGNN successfully maps different messages to different equivariant embeddings.}
    \vspace{-2ex}
    \label{fig:comparison}
\end{figure}

\subsubsection{\textbf{Comparison with existing GNNs}}
Figure~\ref{fig:comparison} compares the equivariant message functions of non-equivariant GNN, EGNN, and DEGNN. Since EGNN is equivariant to rotations, it will map all isometric interactions to equivariant embeddings. And without equivariant constraints, GNN tends to map all distinct interactions to different embeddings. Thus, they all fail to distinguish interactions that lead to different trajectories in Figure~\ref{fig:comparison}. In contrast, DESIGN only maintains the necessary symmetry and successfully learns the message embeddings.

Equivariant inductive biases are crucial to the representation ability of GNNs since they have to maintain symmetry when performing non-linear transformations. For example, EGNN-like models (e.g., GMN~\cite{DBLP:conf/iclr/0001HRX0H22}) rely on the inner product of translation-invariant geometric features (e.g., relative distance, relative velocity, and absolute velocity), which is not able to capture all types of interactions (e.g., $\bm{q}_i\bm{q}_j$ or $\bm{q}_i\bm{\dot{q}}_j$). As illustrated in Eq.~\ref{eq:point}, by relaxing the equivariant group from $E(n)$ to point groups, DEGNN can utilize any symmetry-breaking feature set, which is beneficial to approximate the underlying object interaction functions~\cite{cheng2016wide,DBLP:conf/ijcai/GuoTYLH17}.

\begin{figure}[t]
    \centering
    \includegraphics[width=0.5\textwidth]{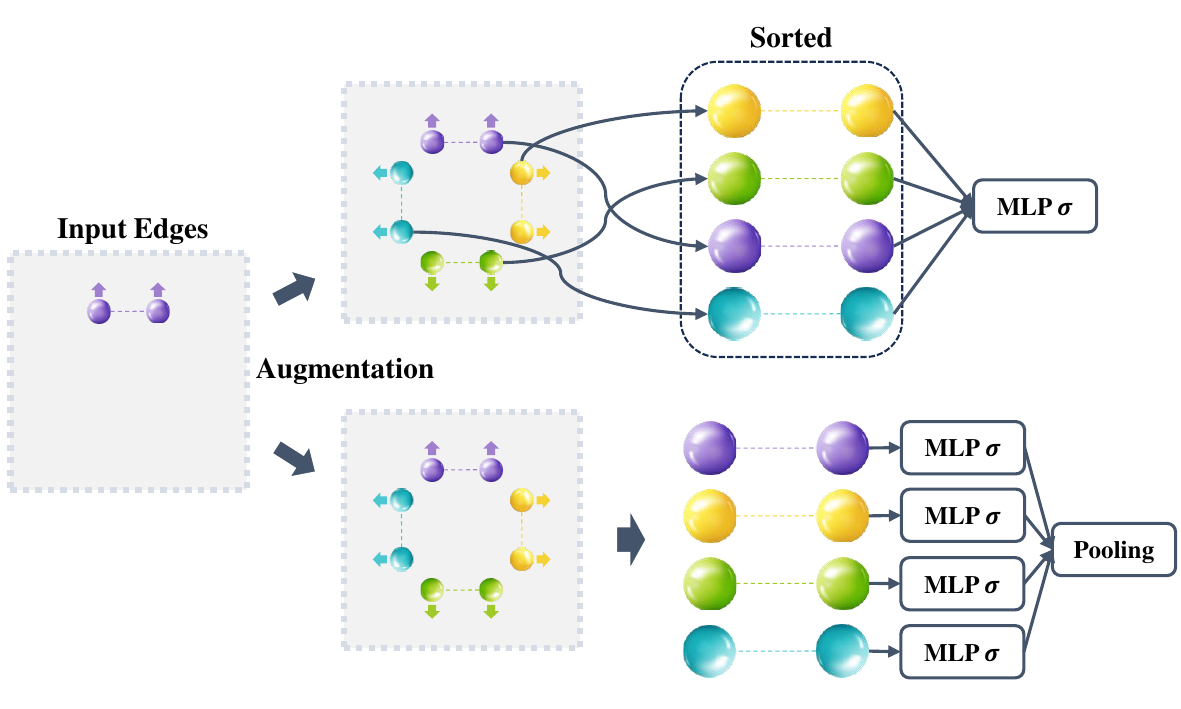}
    \caption{Two realizations of permutation-invariant embedding functions. We omit velocity features in MLP inputs for simplicity. Ranking-based methods rearrange the unordered set to ordered features while pooling-based approaches utilize a permutation-invariant function to aggregate features.}
    \vspace{-4ex}
    \label{fig:realization}
\end{figure}

\subsection{Implementation}
In this section, we present two realizations of the permutation-invariant function $\phi$ - ranking-based and pooling-based. Our objective is to generate a representation of the geometric feature set.
Figure~\ref{fig:realization} illustrates their designs.

\subsubsection{\textbf{Ranking}} 
Since MLP is permutation-sensitive, directly concatenating a set of geometric features as input will produce different outputs when the permutation is changed. To ensure unique embedding for any permutations, we could first sort the features before feeding them into the MLP. Thus, we have:
\begin{equation}
\begin{aligned}
    \phi(\{(\bm{O}\bm{x}, \bm{h})\}_{\bm{O}\in P}) = \sigma(\text{Sort}(\{(\bm{O}\bm{x})\}_{\bm{O}\in P}), \bm{h}).
\end{aligned}\label{eq:sort}
\end{equation}
Here we sort these vectors by the magnitude of each dimension in a descending order. However, the choice of ordering may affect the model performance and manually finding the optimal ordering is challenging and time-consuming.

\subsubsection{\textbf{Permutation-invariant Pooling}}
Since the augmented geometric features could be viewed as independent spatial points, an efficient way to obtain permutation-invariant embedding is through embedding pooling. We first generate embeddings of each point and then employ a permutation-invariant function to aggregate their embeddings. Specifically, the pooling function is defined as follows:
\vspace{-2ex}
\begin{equation}
\begin{aligned}
    \phi(\{(\bm{O}\bm{x}, \bm{h})\}_{\bm{O}\in P}) = \text{Aggregation}(\{\sigma((\bm{O}\bm{x}), \bm{h})\}_{\bm{O}\in P}),
\end{aligned}
\end{equation}
where $\sigma$ is an MLP and the common choices of the aggregation function are mean or sum.

To better describe the symmetry environment in practice, we further propose to employ self-attention to aggregate these geometric embeddings. Let $\bm{e}_i$ denotes the edge embedding of transformation $\bm{O}_i\in P$. Then the computation of message embedding $\bm{m}$ is:
\begin{equation}
\begin{aligned}
\bm{m} &= \sum_{j=1}^{d}\alpha_{ij}\sigma_v(\bm{e}_{j}),\\
\alpha_{ij} &= \frac{\text{exp}(<\sigma_q (\bm{e}_{i}), \sigma_k (\bm{e}_{j})>)}{\sum_{j=1}^{d}\text{exp}(<\sigma_q (\bm{e}_{i}), \sigma_k (\bm{e}_{j})>)},
\end{aligned}\label{eq:attn}
\end{equation}
where $<\cdot,\cdot>$ denotes the inner product of two vectors. $\sigma_q, \sigma_k, \sigma_v$ are MLPs and $\alpha_{ij}$ represents the attention weight. 

\begin{proposition}\label{thm:prop}
The self-attention pooling function is permutation-invariant.
\end{proposition}
\noindent
The proof is reported in the Appendix~\ref{proof:prop} where we show that changing the input permutation will not lead to different message embedding.

\section{Experiments}
In this section, we extensively evaluate our methods on micro-level and macro-level dynamic systems to answer the following research questions:
\begin{itemize}
    \item \textbf{RQ1}: How does the proposed model perform on different types of physical systems compared to the state-of-the-art baselines?
    \item \textbf{RQ2}: Can our model learn the discrete symmetry of the complex real-world dynamics?
    \item \textbf{RQ3}: What are the effects of discrete equivariant message-passing components and graph pooling strategies?
    \item \textbf{RQ4}: How powerful is our method when we vary training size and target interval?
    
\end{itemize}

\subsection{Settings}

\subsubsection{\textbf{Datasets}}
We evaluate model performance on diverse micro-level and macro-level physical dynamics. The micro-level systems are particles and molecules:
\begin{itemize}
    \item \textbf{Particle}: Two Simulated 5-body systems driven by Coulomb or gravitational forces. To evaluate the model performance on particle dynamics of different symmetries, we generate particle trajectories within the cube, square prism, and rectangle prism boundary, which are described by different point groups. We randomly generate 200/2000/2000 trajectories for training/validation/testing.
    \item \textbf{Molecule}: We further evaluate our model on two complex and large molecular simulation datasets~\cite{batzner20223} - LiPS and Li\textsubscript{4}P\textsubscript{2}O\textsubscript{7} contain 83 and 208 atoms within a rectangular prism and parallelepiped boundaries, respectively. 2000/2000/2000 frames are randomly selected for training/validation/testing.
\end{itemize}

Besides, we conduct experiments on macro-level agent dynamics as well. For these datasets, 70\%/10\%/20\% frames are randomly selected for training/validation/testing. Their details are as follows:
\begin{itemize}
    \item \textbf{Crowd}: We use public bidirectional crowd dynamic data that are built up by the Institute for Advanced Simulation 7:Civil Safety Research of ForschungszentrumJülich~\cite{cao2017fundamental}. Pedestrians in these scenarios exhibit a high frequency of interactions. Based on whether they form social groups and the pedestrian number of starting positions, we classify the dataset into low-imbalanced individuals, low-imbalanced groups, high-imbalanced individuals, and high-imbalanced groups.

    \item \textbf{Vehicle}: The vehicle dynamics are obtained from HighD~\cite{highDdataset} dataset, which is recorded on multi-lane highway. The dataset includes 110,500 vehicles covering a total driving distance of 44,500 kilometers and a total of 147 driving hours. Besides, it comprises a total of six distinct highway scenarios and we conducted experiments for each scenario to validate the effectiveness of our model. 
\end{itemize}

The dataset statistics are shown in Appendix  Table~\ref{table:dataset_appendix_1} and Table~\ref{table:dataset_appendix_2}. More details on N-body system generation, molecule data, and the figure illustration of macro-level systems are shown in Appendix~\ref{sec:details}.

\subsubsection{\textbf{Baselines}}
We compare DEGNN against various baselines: (1) non-equivariant GNN; (2) equivariant GNNs: Radial Field~\cite{kohler2019equivariant}, EGNN~\cite{satorras2021n}, GMN~\cite{DBLP:conf/iclr/0001HRX0H22}, EqMotion~\cite{xu2023eqmotion}; (3) in addition, for particle and molecular systems, we further compare DEGNN with TFN~\cite{thomas2018tensor}, SE(3) Transformer~\cite{fuchs2020se}, and SAKE~\cite{DBLP:conf/iclr/WangC23}; (4) for crowd and vehicle, we further compare DEGNN with trajectory forecasting models S-LSTM~\cite{DBLP:conf/cvpr/AlahiGRRLS16} and TransF~\cite{giuliari2021transformer}, and graph simulation models GNS~\cite{DBLP:conf/icml/Sanchez-Gonzalez20}, and CrowdSim~\cite{DBLP:conf/www/ShiYL23}.

\subsubsection{\textbf{Implementation details}}
we empirically find that the following hyper-parameters generally work well, and use them across all experiments: Batch size 100, the hidden dimension 64, weight decay $1\times 10^{-12}$. All models are set to four layers. Due to insufficient GPU memory, the batch size for the EqMotion model is set to 50 for the LiPS dataset and 5 for the Li\textsubscript{4}P\textsubscript{2}O\textsubscript{7} dataset.
The learning rates of the micro-level and macro-level datasets are 0.0003 and 0.0005. All models are trained for 5000 epochs with an early stopping strategy of 100. We use Mean Square Error (MSE) as our metric to measure the loss between prediction and the ground truth. The point group used in DEGNN can be found in Appendix~\ref{table:group_appendix}. All models are implemented based on Pytorch and PyG library~\cite{fey2019fast}, trained on GeForce RTX 4090 GPU.

\begin{table*}[t]
\setlength{\tabcolsep}{1mm}
\centering 
\caption{
MSE of all models on micro-level systems. Bold font indicates the best result and Underline is the strongest baseline. We report both mean and standard deviation that are computed over 5 runs. “OOM” denotes out of memory. The official implementation of Eqmotion report "Nan" on Li\textsubscript{4}P\textsubscript{2}O\textsubscript{7} dataset. 
In particle systems, (a,b,c) denotes the side length of the boundary, where (5,5,5) and (4,4,4) are cube, (5,4,4) and (5,4,3) are square prism and rectangle prism.}
\label{table:overall}
\resizebox{0.9\textwidth}{!}{
\begin{tabular}{c|cccc|cccc|cc}
\toprule
 \multirow{2}{*}{\textbf{Model}}  & \multicolumn{4}{c|}{\textbf{Charged Particle} ($\times 10^{-1}$)} & \multicolumn{4}{c|}{\textbf{Gravitational Particle} ($\times 10^{-1}$)} & \multicolumn{2}{c}{\textbf{Molecular}($\times 10^{-1}$)}\\
 & (5,5,5) & (5,4,4) & (5,4,3) & (4,4,4) & (5,5,5) & (5,4,4) & (5,4,3) & (4,4,4) & LiPS & Li\textsubscript{4}P\textsubscript{2}O\textsubscript{7}\\
\midrule

   \textbf{GNN} & 3.12{$\pm$ 0.49} & 4.46{$\pm$ 0.15} & 3.24{$\pm$ 0.13} & 3.14{$\pm$0.19 } & 3.81{$\pm$ 0.54} & \underline{3.32{$\pm$ 0.37}}  & \underline{3.21{$\pm$ 0.27}} & \underline{3.20{$\pm$ 0.61}} & \underline{4.39{$\pm$ 0.71}}  & \textbf{8.59{$\pm$ 0.51}}\\
   \textbf{TFN} & 5.03{$\pm$ 0.33} & 5.53{$\pm$ 0.65} & 6.15{$\pm$ 0.30} & 5.53{$\pm$0.65 } &  6.70{$\pm$ 0.17} & 7.49{$\pm$ 0.24}  & 8.07{$\pm$ 0.55} & 7.74{$\pm$ 0.32} & OOM  & OOM\\
   \textbf{Radial Field} & 1.94{$\pm$ 0.03} & 2.58{$\pm$ 0.07} & 3.04{$\pm$ 0.01} & 2.77{$\pm$ 0.10} &  4.76{$\pm$ 0.03} & 5.51{$\pm$ 0.06} & 5.85{$\pm$ 0.12} & 5.29{$\pm$ 0.20} & 10.0{$\pm$ 0.06}  & 13.3{$\pm$ 0.02}\\
   \textbf{SE(3)-Tr.} & 4.65{$\pm$ 0.08} & 5.15{$\pm$ 0.01} & 5.70{$\pm$ 0.10} & 5.29{$\pm$ 0.12} &  9.03{$\pm$ 0.20} & 10.1{$\pm$ 0.27}  & 10.54{$\pm$ 0.23} & 10.3{$\pm$ 0.30} & OOM  & OOM\\

   \textbf{EGNN} & 4.92{$\pm$ 0.12} & 5.62{$\pm$ 0.03} & 6.33{$\pm$ 0.05} & 5.67{$\pm$ 0.02} &  10.3{$\pm$ 0.04} & 11.5{$\pm$ 0.03}  & 12.8{$\pm$ 0.06} & 11.5{$\pm$ 0.04} & 7.97{$\pm$ 0.49}  & 13.8{$\pm$ 0.04}\\
   \textbf{GMN} & \underline{1.82{$\pm$ 0.13}} & \underline{2.23{$\pm$ 0.03}} & \underline{2.64{$\pm$ 0.07}} & \underline{2.41{$\pm$ 0.05}} &  \underline{3.78{$\pm$ 0.15}} & 4.48{$\pm$ 0.13} & 5.07{$\pm$ 0.10} & 4.44{$\pm$ 0.20} & 6.26{$\pm$ 0.65} & 13.0{$\pm$ 0.04}\\
   \textbf{SAKE} & 4.22{$\pm$ 2.42} & 3.88{$\pm$ 0.55} & 3.87{$\pm$ 0.05} & 3.80{$\pm$ 0.10} &  5.88{$\pm$ 0.33} & 7.21{$\pm$ 0.34} & 7.60{$\pm$ 0.37} & 6.89{$\pm$ 0.40} & 11.5{$\pm$ 9.76}  & 12.8{$\pm$ 0.27}\\
   \textbf{EqMotion} & 2.14{$\pm$ 0.24} & 2.70{$\pm$ 0.34} & 3.27{$\pm$ 0.25} & 2.96{$\pm$ 0.34} &  4.23{$\pm$ 0.24} & 4.95{$\pm$ 0.18} & 5.62{$\pm$ 0.18} & 4.90{$\pm$ 0.19} & 15.2{$\pm$ 9.8}  & Nan\\

\midrule
\textbf{DEGNN} & \textbf{1.23{$\pm$ 0.06}} & \textbf{1.43{$\pm$ 0.07}}  &  \textbf{1.24{$\pm$ 0.05}} & \textbf{1.54{$\pm$ 0.07}}  & \textbf{1.46{$\pm$ 0.17}} & \textbf{1.73{$\pm$ 0.21}} & \textbf{1.56{$\pm$ 0.09}} & \textbf{2.02{$\pm$ 0.18}} & \textbf{4.36{$\pm$ 0.31}} & \underline{9.86{$\pm$ 0.97}}\\

\bottomrule
\end{tabular}
}
\vspace{-1ex}
\end{table*}

\begin{table*}[t]
\setlength{\tabcolsep}{0.5mm}
\centering 
\vspace{-1ex}
\caption{
MSE of all models on macro-level systems. Bold font indicates the best result and Underline is the strongest baseline. We report both mean and standard deviation that are computed over 5 runs.
}
\label{table:macro}
\resizebox{0.9\textwidth}{!}{
\begin{tabular}{c|cccc|cccccc}
\toprule
 \multirow{2}{*}{\textbf{Model}}  & \multicolumn{4}{c|}{\textbf{Crowd} ($\times 10^{-3}$)} & \multicolumn{6}{c}{\textbf{Vehicle} ($\times 10^{-2}$)} \\
 & Ind.-Low & Ind.-High & Group-Low & Group-High & Highway 1 & Highway 2 & Highway 3 & Highway 4 & Highway 5 & Highway 6\\
\midrule

   \textbf{GNN}  & 1.43{$\pm$ 0.25}& \underline{0.86{$\pm$0.24}} & \underline{0.8{$\pm$ 0.28}}& 1.20{$\pm$ 0.10} & \underline{0.61{$\pm$ 0.28}} & \underline{0.20{$\pm$ 0.07}} & \underline{0.67{$\pm$ 0.29}} & \underline{1.02{$\pm$ 0.77}} & \underline{0.39{$\pm$ 0.19}} & 2.27{$\pm$ 1.27} \\
  \textbf{S-LSTM} & 3.43{$\pm$ 0.01}& 1.89{$\pm$0.09}  & 3.83{$\pm$ 1.25}& 2.57{$\pm$0.49} & 1.66{$\pm$ 0.05}&0.49{$\pm$0.03}  &0.78{$\pm$ 0.03}&1.72{$\pm$0.01}&1.22{$\pm$ 0.05}&1.23{$\pm$0.07}\\
   \textbf{Radial Field} & 3.21{$\pm$ 1.16}& 2.43{$\pm$0.17} &3.76{$\pm$ 0.45}& 3.39{$\pm$0.87}  & 512{$\pm$ 395} & 303{$\pm$ 270} & 180{$\pm$ 127} & 44.0{$\pm$ 34.9} & 477{$\pm$ 301} & 240{$\pm$ 211} \\
   \textbf{GNS} & 3.7{$\pm$ 0.49}& 4.14{$\pm$1.02} & 5.25{$\pm$ 1.61}& 3.66{$\pm$0.71}  & 1.58{$\pm$ 0.78} & 3.00{$\pm$ 2.46} & 1.54{$\pm$ 0.55} & 5.13{$\pm$ 2.07} & 4.53{$\pm$ 5.03} & 2.43{$\pm$ 0.40} \\
  \textbf{TransF} & 1.59{$\pm$0.07}& 1.76{$\pm$0.04} & 2.72{$\pm$0.04}& 1.89{$\pm$0.04}  & 1.61{$\pm$0.87}& 0.59{$\pm$0.08} & 0.70{$\pm$0.25}& 1.36{$\pm$0.70}& 0.68{$\pm$0.30}& \underline{0.94{$\pm$0.34}}\\
   \textbf{EGNN} & 5.94{$\pm$ 1.05}& 2.17{$\pm$0.84} & 2.02{$\pm$ 0.32}& 1.48{$\pm$ 0.30}  & 24.9{$\pm$ 7.79} & 52.4{$\pm$ 27.4} & 42.6{$\pm$ 8.25} & 19.3{$\pm$ 6.38} & 46.1{$\pm$ 4.84} & 35.8{$\pm$ 23.8} \\
      \textbf{GMN} & \underline{0.44{$\pm$ 0.07}}& 1.17{$\pm$0.05} & 0.91{$\pm$ 0.36}& \underline{0.71{$\pm$ 0.48}}  & 1.41{$\pm$ 0.76} & 0.74{$\pm$ 0.27} & 1.30{$\pm$ 0.45} & 2.48{$\pm$ 0.90} & 2.94{$\pm$ 2.00}   & 5.20{$\pm$ 2.99} \\
   \textbf{CrowdSim} & 3.60{$\pm$ 1.25}& 3.14{$\pm$0.95} & 5.77{$\pm$ 0.90}  & 2.93{$\pm$ 1.11}& 1.85{$\pm$ 0.67} & 2.39{$\pm$ 0.36} & 1.25{$\pm$ 0.45} & 11.3{$\pm$ 12.5} & 13.2{$\pm$ 24.3} & 3.52{$\pm$ 1.26}  \\
   \textbf{EqMotion} & 8.00{$\pm$ 7.34}& 5.98{$\pm$1.22} & 6.27{$\pm$ 4.82} & 2.91{$\pm$ 2.54} & 29.8{$\pm$ 23.6} & 1.18{$\pm$ 2.01} & 39.4{$\pm$ 44.0} & 3.77{$\pm$ 7.3} & 541{$\pm$ 432} & 144{$\pm$ 247} \\

\midrule
\textbf{DEGNN} & \textbf{0.32{$\pm$ }0.04} & \textbf{0.41{$\pm$0.04}} & \textbf{0.71{$\pm$0.13}} & \textbf{0.36{$\pm$ 0.04}}  & \textbf{0.07{$\pm$ 0.01}} & \textbf{0.04{$\pm$ 0.07}}  &  \textbf{0.35{$\pm$ 0.35}} & \textbf{0.56{$\pm$ 0.08}} & \textbf{0.11{$\pm$ 0.02}}  &  \textbf{0.27{$\pm$ 0.04}}\\

\bottomrule
\end{tabular}
}
\end{table*}

\subsection{Overall Performance Comparison (RQ1)}
In this section, we validate the effectiveness of our model by comparing it to the baselines.

\subsubsection{\textbf{Micro-level}}
 Table~\ref{table:overall} presents the comparison of different dynamic modeling methods on particle and molecule systems. Based on the results, we have the following observations:
\begin{itemize}
    \item DEGNN outperforms all baselines on all particle datasets in a large gap. Compared with the best baseline, the average enhancements on the Charged and Gravity datasets are 0.9 and 1.7 respectively. For molecular systems, our model surpasses the performance of other models on the LiPS dataset while achieving comparable performance to the best baseline on the Li\textsubscript{4}P\textsubscript{2}O\textsubscript{7} dataset.
    \item  The basic GNN performs better than several equivariant GNNs (e.g., TFN and EGNN), especially in Molecular systems. These results show that $E(n)$-equivariant constraint is too strong to model these dynamics, leading to a decrease in performance.
    \item In particle systems, as the boundary changes from the cube (5,5,5), square prim (5,4,4), to rectangle prism (5,4,3), the degree of discrete symmetry gradually decreases, indicating the size of the corresponding point group is reduced. In these systems, we can observe that the error of most models rises when the symmetry weakens. For instance, in Charged particle systems, the error of Emotion in (5,5,5), (5,4,4), and (5,4,3) boundaries are 2.14, 2.70, 3.27, respectively. In contrast, DEGNN maintains low error in different systems.
    \item Note that dynamics with small boundaries will have more boundary effects. By comparing the results on (5,5,5) and (4,4,4) systems, we can observe that almost all models achieve larger errors in (4,4,4) systems, indicating these dynamics are more complex. DEGNN still achieves the best performance in such cases.

\end{itemize}
\subsubsection{\textbf{Macro-level}}
 Table~\ref{table:macro} depicts the overall results on real-world crowd and vehicle datasets. The observations are as follows: 
\begin{itemize}
    \item GNN outperforms all baselines in 7 of 10 scenarios, indicating incorporating continuous equivariance does not enhance model performance in most cases. On the contrary, DEGNN outperforms all baseline models, with the lowest errors on crowd and vehicle datasets. Such improvement verifies the effectiveness of encoding the discrete symmetry in the dynamic modeling of agents.

    \item Another discovery is that our model DEGNN exhibits a lower standard deviation compared to other equivariant NN models. In the Crowd dataset, DEGNN demonstrates better stability (0.06 average standard deviation) compared to EGNN (0.6), GMN (0.24), and EqMotion (4).
\end{itemize}

\subsection{Generalization Experiments (RQ2)}\label{sec:rq2}
In this section, we conduct two additional experiments across different directions and scenarios on vehicle datasets to validate the generalization ability of DEGNN.
\subsubsection{\textbf{Different directions}}

\begin{figure}[t]
    \centering
    \includegraphics[width=0.45\textwidth]{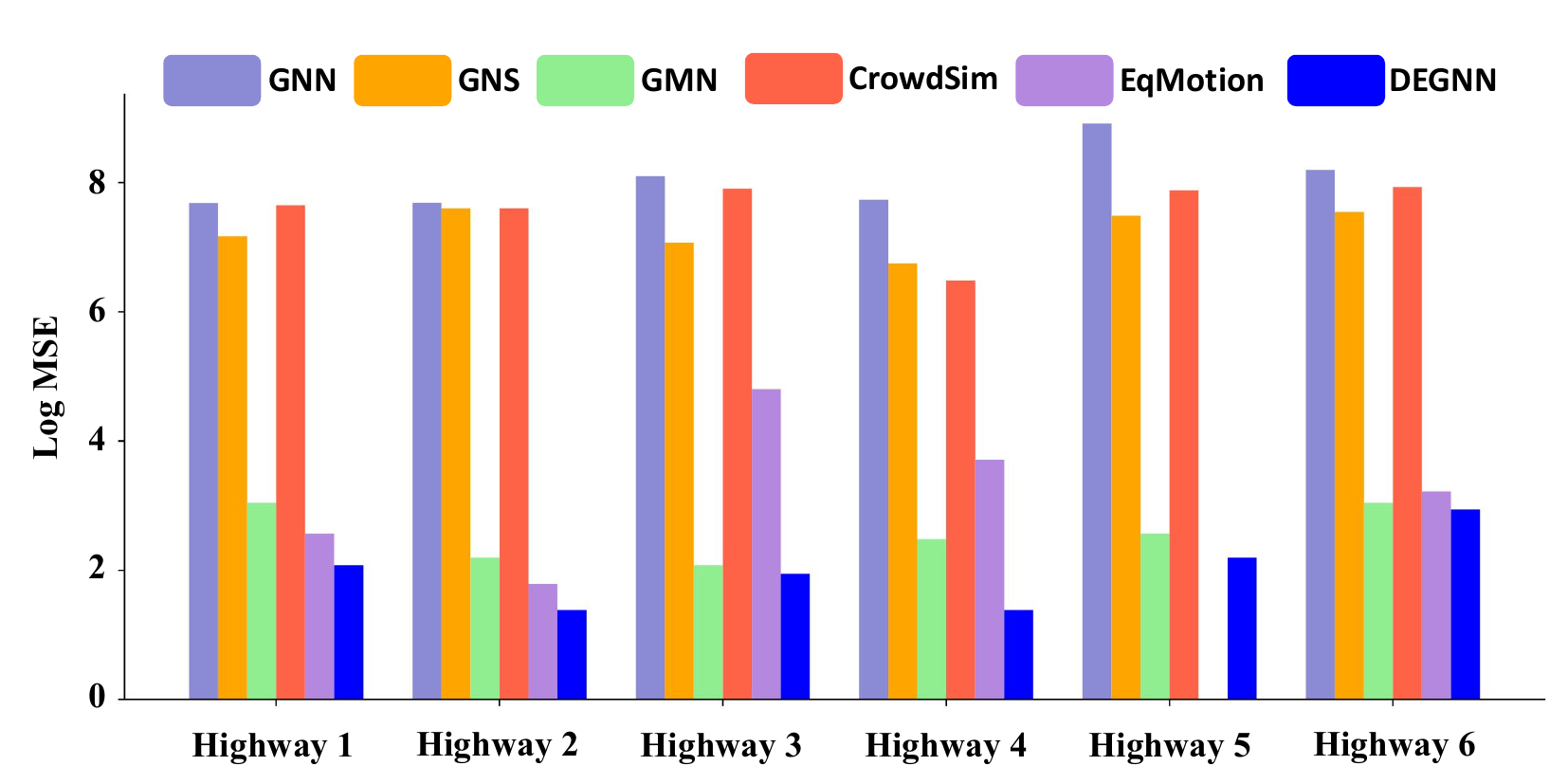}
    \vspace{-1ex}
    \caption{MSE of models on vehicle dataset. They are trained on trajectories from the left to right and tested on those from the right to left. (Official EqMotion reports "Nan" in Highway 5 datasets; \bm{$Log\, MSE = log (100 \times MSE + 1)$})}
    \vspace{-2ex}
    \label{fig:bar}
\end{figure}

In each scenario, models are trained on trajectories in one direction (from left to right) and evaluated on trajectories in the opposite direction (from right to left). We compare DEGNN with Eqmotion, GNN, GMN, GNS, and CrowdSim.
Figure~\ref{fig:bar} displays the result. We can find that without observing the opposite direction data, GNN, GNS, and CrowdSim achieve high errors, indicating their weak generalization ability in dynamic directions.
$E(n)$-equivariant models exhibit much stronger generalization than the above models and DEGNN outperforms GMN and EqMotion by a significant margin, demonstrating the discrete equivariance achieves the best generalization performance.

\subsubsection{\textbf{Cross scenarios}}
\begin{figure}[t]
    \centering
    \includegraphics[width=0.35\textwidth]{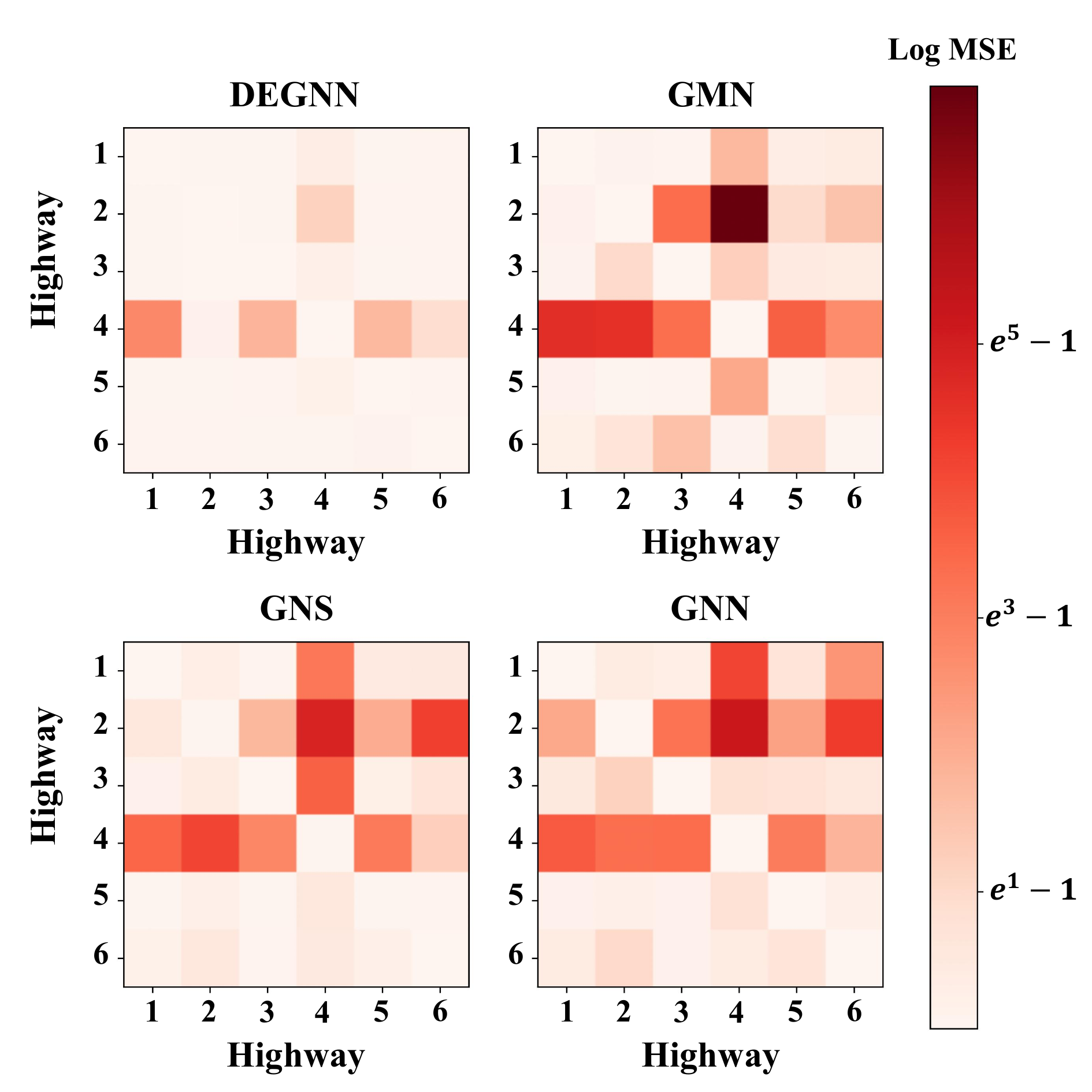}
    \vspace{-1ex}
    \caption{Generalization experiments across different scenarios on vehicle dynamic datasets. Row/Column denotes the training/testing scenarios. (\bm{$Log\, MSE = log (MSE + 1)$})}
    \vspace{-3ex}
    \label{fig:heat}
\end{figure}

Models are trained on trajectories from one scenario and evaluated on trajectories from other scenarios. We present the results of GNN, DEGNN, GMN, and GNS in Figure~\ref{fig:heat} where a lighter color indicates a lower error. From the figure, we can observe that all compared methods achieve high generalization errors in multiple cases, indicating they overfit the training data instead of learning the real interaction patterns. Compared to these methods, DEGNN demonstrates significantly better cross-scenario generalization capabilities.

\subsection{Ablation Study (RQ3)}

\begin{table*}[t]
\setlength{\tabcolsep}{4mm}
\centering 
\caption{
Ablation Studies on model designs. Bold font indicates the best result. We report both mean and standard deviation that are computed over 5 runs. In molecular systems, (a,b,c) denotes the side length of the boundary, where (5,5,5) and (4,4,4) are cube, (5,4,4) and (5,4,3) are square prism and rectangle prism.
}
\label{table:ablation}
\resizebox{0.9\textwidth}{!}{
\begin{tabular}{c|cccc|cccc}
\toprule
 \multirow{2}{*}{\textbf{Model}}  & \multicolumn{4}{c|}{\textbf{Charged Particle} ($\times 10^{-1}$)} & \multicolumn{4}{c}{\textbf{Gravitational Particle} ($\times 10^{-1}$)} \\
 & (5,5,5) & (5,4,4) & (5,4,3) & (4,4,4) & (5,5,5) & (5,4,4) & (5,4,3) & (4,4,4)\\
\midrule
  \textbf{E(n)}  & 1.78{$\pm$0.15} & 2.12{$\pm$0.14}& 2.01{$\pm$ 0.06} & 2.36{$\pm$0.13}& 2.32{$\pm$0.05} & 2.83{$\pm$ 0.12} & 3.35{$\pm$0.19 } & 3.29{$\pm$0.17}\\
  \textbf{O(n)} & 2.46{$\pm$0.82} & 2.13{$\pm$0.15}& 1.79{$\pm$ 0.07} & 2.03{$\pm$0.26}& 1.73{$\pm$0.13} & 3.49{$\pm$ 0.92} & 2.27{$\pm$0.08 } & 2.52{$\pm$0.48}\\
  \textbf{T(n)} & 2.79{$\pm$0.04} & 2.80{$\pm$ 0.13}& 2.18{$\pm$ 0.07} & 3.76{$\pm$0.06}& 5.98{$\pm$0.68} & 3.57{$\pm$ 0.24} & 3.35{$\pm$0.09 } & 6.63{$\pm$0.08}\\
  \midrule
    \textbf{Ranking} & 1.95{$\pm$ 0.22} & 1.63{$\pm$0.10}& 1.51{$\pm$0.07}& 2.11{$\pm$0.18}& 2.06{$\pm$0.09} & 1.79{$\pm$ 0.10} & 1.75{$\pm$ 0.08} & 2.36{$\pm$0.11} \\
  \textbf{Sum Pooling} & 25.3{$\pm$ 12.2} & 2.47{$\pm$0.64} & 1.82{$\pm$ 0.09} &13.0{$\pm$ 5.15} & 13.55{$\pm$4.77} & 4.16{$\pm$ 0.67} & 3.21{$\pm$ 0.53} & 52.9{$\pm$61.4}\\
   \textbf{Mean Pooling} & 2.09{$\pm$ 0.23} & 2.39{$\pm$ 0.12} &1.52{$\pm$ 0.14} & 2.10{$\pm$ 0.11} & 2.13{$\pm$0.42} & 2.10{$\pm$ 0.24} & 1.78{$\pm$ 0.19} & 2.60{$\pm$0.24} \\

\midrule
\textbf{Default} & \textbf{1.23{$\pm$ 0.06}} & \textbf{1.43{$\pm$ 0.07}}  &  \textbf{1.24{$\pm$ 0.05}} & \textbf{1.54{$\pm$ 0.07}} & \textbf{1.46{$\pm$ 0.17}} & \textbf{1.73{$\pm$ 0.21}} & \textbf{1.56{$\pm$ 0.09}} & \textbf{2.02{$\pm$ 0.18}}\\

\bottomrule
\end{tabular}
}
\end{table*}

To illustrate the effectiveness of each model deisgn to the overall performance of our model, we compare the default settings of DEGNN with three model feature variants by replacing the feature in Eq.~\ref{eq:point} with : (1) \textbf{E(n)}: $\bm{x}=(||\bm{q}_{i}-\bm{q}_{j}||^2)$, which satisfied rotation, translation, and reflection equivariance; (2) \textbf{O(n)}: $\bm{x}=(||\bm{q}_{i}||^2, ||\bm{q}_{j}||^2)$, which satisfied rotation and reflection equivariance; (3) \textbf{T(n)}: $\bm{x}=(\bm{q}_{i}- \bm{q}_{j})$, which satisfied translation equivariance, and three permutation invariant function variants: (4) \textbf{Ranking}: sort the feature vector based on each dimension of geometric vectors; (5) \textbf{Sum Pooling}: sum the feature embedding; (6) \textbf{Mean Pooling}: averaging the feature embedding; Table~\ref{table:ablation} shows the results on particle systems. The results of the other systems are shown in Table~\ref{table:ablation_appendix}.

\subsubsection{\textbf{Effect of discrete equivariant message passing}}
 We can find that variant \textbf{T(n)} exhibits the highest error among the three variants. For example, the average loss for \textbf{E(n)}, \textbf{O(n)} and \textbf{T(n)} in the Charged dataset is 2.07, 2.10, and 2.88 respectively. This discrepancy is mainly attributed to the fact that, in the presence of boundaries, the dynamic is not equivariant to translation. 
 DEGNN outperforms variants in all cases, demonstrating the effectiveness of discrete equivariance.

\subsubsection{\textbf{Effect of graph pooling strategies}}
 Among the three implementations of DEGNN, sum pooling demonstrates the lowest model performance. In scenarios with high symmetry, such as Charged (5,5,5) and Charged (4,4,4), the losses reach 25 and 13, respectively. The reason is that the size of $O_h$ (i.e., the point group of Cube) is large which makes the sum pooling unstable. On the contrary, the self-attention pooling is stable and consistently outperforms other variants.

\subsection{Sensitivity Analysis (RQ4)}

\begin{figure*}[t]
    \centering
    \includegraphics[width=0.9\textwidth]{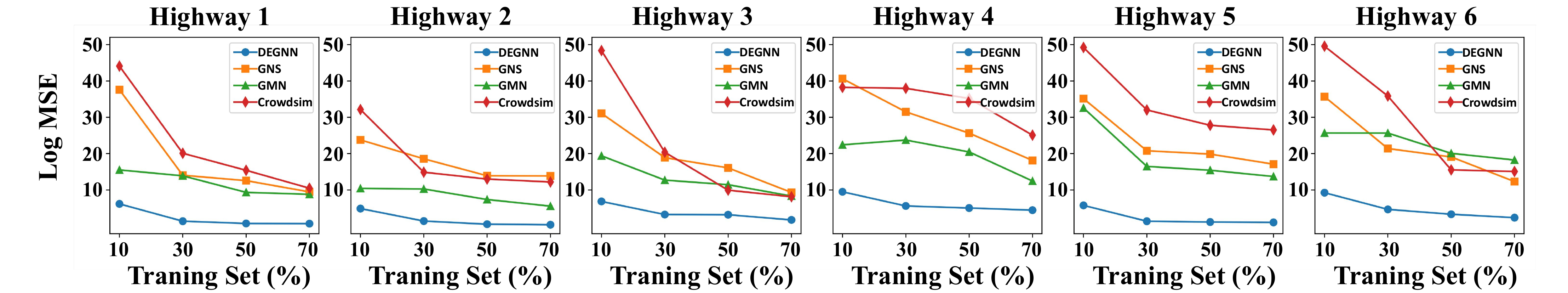}
    \includegraphics[width=0.9\textwidth]{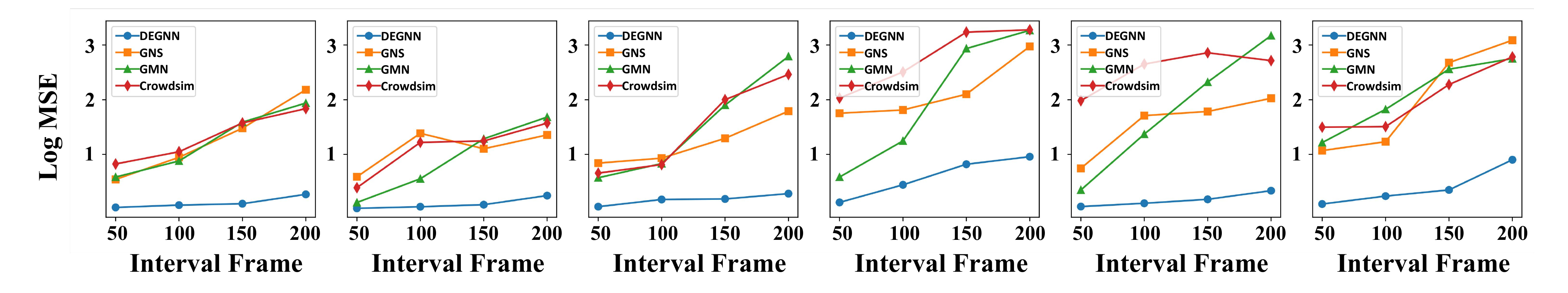}
    \caption{The rollout-MSE curves on 6 vehicle scenarios. DEGNN achieves the best performance.(\bm{$Log\, MSE = log (100 \times MSE + 1)$})}
    \vspace{-3ex}
    \label{fig:train}
\end{figure*}

In this section, we conducted experiments on vehicle datasets by varying the training set size and target time interval.

\subsubsection{\textbf{Performance w.r.t. training size}}
The first row of Figure~\ref{fig:train} illustrates the results of leveraging 10\%, 30\%, 50\%, and 70\% of the dataset for training.
As the training dataset size decreases to 10\%, the loss of DEGNN consistently remains below 10, while other models may increase to 40 or even 50. Such results show that DEGNN is data efficient.

\subsubsection{\textbf{Performance w.r.t. target interval}}
Figure~\ref{fig:train} depicts how the model loss changes with varying prediction frame intervals, ranging from 50 frames (2s) to 200 (8s) frames. With the increment in interval frames, DEGNN consistently sustains a loss below 1, whereas other models might escalate to 2 or 3. Such results demonstrate that DEGNN can generalize to long-term predictions.

\section{Related work}
In this section, we review the most related works from the perspective of physical systems.

\paragraph{\textbf{Particle and Molecule}}
Graph neural networks~\cite{liu2020modelling} have shown promising performance in learning complex dynamics of N-body systems (e.g., particles and molecules). IN~\citep{battaglia2016interaction}, NRI~\citep{kipf2018neural}, and HRN~\citep{mrowca2018flexible} are pioneer works that model physical objects and relations as graphs and learn their interaction and evolution. Recent studies have considered the underlying physical symmetry of systems. TFN~\citep{thomas2018tensor}, SE(3) Transformer~\citep{fuchs2020se}, and SEGNN~\citep{DBLP:conf/iclr/BrandstetterHPB22} utilize spherical harmonics to construct rotation equivariant models with higher-order geometric representations. 
Several works~\cite{finzi2020generalizing,hutchinson2021lietransformer} leverage the Lie convolution to extend equivariance on Lie groups.
In addition to these methods, a series of studies~\citep{satorras2021n,DBLP:conf/iclr/WangC23,DBLP:conf/iclr/0001HRX0H22,xu2023eqmotion,wu2024equivariant} apply scalarization techniques to introduce equivariance into the message-passing process in GNNs. Nevertheless, these methods incorporate continuous equivariant constraints, which are too strong for various physical systems.

\paragraph{\textbf{Crowd}}
Traditional crowd dynamic modeling models include rule-based~\cite{reynolds1987flocks}, force-based~\cite{helbing1995social,saboia2012crowd}, and velocity-based models~\cite{fiorini1998motion}. Nevertheless, the designed formulas are insufficient to model complex and uncertain real-world dynamics. With the development of deep learning, recent studies~\cite{DBLP:conf/cikm/YuZLJ23,DBLP:conf/cikm/YuZLJ23} seek neural solutions to improve model performance. For example, CrowdSim~\cite{DBLP:conf/www/ShiYL23} learns crowd trajectories under the framework of GNS~\cite{DBLP:conf/icml/Sanchez-Gonzalez20}. However, these methods ignore the symmetry of crowd dynamics, leading to suboptimal generalization ability.

\paragraph{\textbf{Vehicle}}
Current methods for vehicle trajectory prediction are mainly based on sequence models such as recurrent models~\cite{DBLP:conf/eccv/SalzmannICP20,DBLP:conf/cvpr/DeoT18,DBLP:conf/ivs/DeoT18} and transformers~\cite{giuliari2021transformer,DBLP:conf/iccv/0007WOK21}. ECCO~\cite{DBLP:conf/iclr/WaltersLY21} embeds the rotation symmetries in continuous convolution to improve data efficiency and make physically consistent predictions. However, it still follows the continuous equivariance, which constrains its representation ability.

\section{Conclusion and Future work}
In this work, we propose a general DEGNN framework for learning micro-level and macro-level physical dynamics in boundary environments. It is proved to be equivariant to the given point group via data augmentation and permutation-invariant embedding functions. We propose ranking-based and pooling-based implementations of DEGNN. Extensive experiments conducted on various physical dynamics show that DEGNN outperforms all competing methods. Generalization experiments and sensitivity analysis demonstrate the superior generalization ability of our model. Ablation studies further validate the effectiveness of our model designs. In the future, we are interested in (1) extending our work to solve multiple physical dynamics jointly; (2) evaluating DEGNN on multi-step trajectories; (3) learning a ranking function for ranking-based DEGNN.

\begin{acks}
This work was supported by NSFC Grant No. 62206067, HKUST-HKUST(GZ) 20 for 20 Cross-campus Collaborative Research Scheme C019 and Guangzhou-HKUST(GZ) Joint Funding Scheme 2023A03J0673.
\end{acks}

\bibliographystyle{ACM-Reference-Format}
\bibliography{sample-base}


\begin{thebibliography}{58}


\ifx \showCODEN    \undefined \def \showCODEN     #1{\unskip}     \fi
\ifx \showDOI      \undefined \def \showDOI       #1{#1}\fi
\ifx \showISBNx    \undefined \def \showISBNx     #1{\unskip}     \fi
\ifx \showISBNxiii \undefined \def \showISBNxiii  #1{\unskip}     \fi
\ifx \showISSN     \undefined \def \showISSN      #1{\unskip}     \fi
\ifx \showLCCN     \undefined \def \showLCCN      #1{\unskip}     \fi
\ifx \shownote     \undefined \def \shownote      #1{#1}          \fi
\ifx \showarticletitle \undefined \def \showarticletitle #1{#1}   \fi
\ifx \showURL      \undefined \def \showURL       {\relax}        \fi
\providecommand\bibfield[2]{#2}
\providecommand\bibinfo[2]{#2}
\providecommand\natexlab[1]{#1}
\providecommand\showeprint[2][]{arXiv:#2}

\bibitem[Alahi et~al\mbox{.}(2016)]%
        {DBLP:conf/cvpr/AlahiGRRLS16}
\bibfield{author}{\bibinfo{person}{Alexandre Alahi}, \bibinfo{person}{Kratarth Goel}, \bibinfo{person}{Vignesh Ramanathan}, \bibinfo{person}{Alexandre Robicquet}, \bibinfo{person}{Li Fei{-}Fei}, {and} \bibinfo{person}{Silvio Savarese}.} \bibinfo{year}{2016}\natexlab{}.
\newblock \showarticletitle{Social {LSTM:} Human Trajectory Prediction in Crowded Spaces}. In \bibinfo{booktitle}{\emph{CVPR}}. \bibinfo{pages}{961--971}.
\newblock


\bibitem[Battaglia et~al\mbox{.}(2016)]%
        {battaglia2016interaction}
\bibfield{author}{\bibinfo{person}{Peter Battaglia}, \bibinfo{person}{Razvan Pascanu}, \bibinfo{person}{Matthew Lai}, \bibinfo{person}{Danilo Jimenez~Rezende}, {et~al\mbox{.}}} \bibinfo{year}{2016}\natexlab{}.
\newblock \showarticletitle{Interaction networks for learning about objects, relations and physics}.
\newblock \bibinfo{journal}{\emph{Advances in neural information processing systems}}  \bibinfo{volume}{29} (\bibinfo{year}{2016}).
\newblock


\bibitem[Batzner et~al\mbox{.}(2022)]%
        {batzner20223}
\bibfield{author}{\bibinfo{person}{Simon Batzner}, \bibinfo{person}{Albert Musaelian}, \bibinfo{person}{Lixin Sun}, \bibinfo{person}{Mario Geiger}, \bibinfo{person}{Jonathan~P Mailoa}, \bibinfo{person}{Mordechai Kornbluth}, \bibinfo{person}{Nicola Molinari}, \bibinfo{person}{Tess~E Smidt}, {and} \bibinfo{person}{Boris Kozinsky}.} \bibinfo{year}{2022}\natexlab{}.
\newblock \showarticletitle{E (3)-equivariant graph neural networks for data-efficient and accurate interatomic potentials}.
\newblock \bibinfo{journal}{\emph{Nature communications}} \bibinfo{volume}{13}, \bibinfo{number}{1} (\bibinfo{year}{2022}), \bibinfo{pages}{2453}.
\newblock


\bibitem[Bradley and Cracknell(2010)]%
        {bradley2010mathematical}
\bibfield{author}{\bibinfo{person}{Christopher Bradley} {and} \bibinfo{person}{Arthur Cracknell}.} \bibinfo{year}{2010}\natexlab{}.
\newblock \bibinfo{booktitle}{\emph{The mathematical theory of symmetry in solids: representation theory for point groups and space groups}}.
\newblock \bibinfo{publisher}{Oxford University Press}.
\newblock


\bibitem[Brandstetter et~al\mbox{.}(2022)]%
        {DBLP:conf/iclr/BrandstetterHPB22}
\bibfield{author}{\bibinfo{person}{Johannes Brandstetter}, \bibinfo{person}{Rob Hesselink}, \bibinfo{person}{Elise van~der Pol}, \bibinfo{person}{Erik~J. Bekkers}, {and} \bibinfo{person}{Max Welling}.} \bibinfo{year}{2022}\natexlab{}.
\newblock \showarticletitle{Geometric and Physical Quantities improve {E(3)} Equivariant Message Passing}. In \bibinfo{booktitle}{\emph{ICLR}}.
\newblock


\bibitem[Cao et~al\mbox{.}(2017)]%
        {cao2017fundamental}
\bibfield{author}{\bibinfo{person}{Shuchao Cao}, \bibinfo{person}{Armin Seyfried}, \bibinfo{person}{Jun Zhang}, \bibinfo{person}{Stefan Holl}, {and} \bibinfo{person}{Weiguo Song}.} \bibinfo{year}{2017}\natexlab{}.
\newblock \showarticletitle{Fundamental diagrams for multidirectional pedestrian flows}.
\newblock \bibinfo{journal}{\emph{Journal of Statistical Mechanics: Theory and Experiment}} \bibinfo{volume}{2017}, \bibinfo{number}{3} (\bibinfo{year}{2017}), \bibinfo{pages}{033404}.
\newblock


\bibitem[Cheng et~al\mbox{.}(2016)]%
        {cheng2016wide}
\bibfield{author}{\bibinfo{person}{Heng-Tze Cheng}, \bibinfo{person}{Levent Koc}, \bibinfo{person}{Jeremiah Harmsen}, \bibinfo{person}{Tal Shaked}, \bibinfo{person}{Tushar Chandra}, \bibinfo{person}{Hrishi Aradhye}, \bibinfo{person}{Glen Anderson}, \bibinfo{person}{Greg Corrado}, \bibinfo{person}{Wei Chai}, \bibinfo{person}{Mustafa Ispir}, {et~al\mbox{.}}} \bibinfo{year}{2016}\natexlab{}.
\newblock \showarticletitle{Wide \& deep learning for recommender systems}. In \bibinfo{booktitle}{\emph{DLRS@RecSys}}. \bibinfo{pages}{7--10}.
\newblock


\bibitem[Deo and Trivedi(2018a)]%
        {DBLP:conf/cvpr/DeoT18}
\bibfield{author}{\bibinfo{person}{Nachiket Deo} {and} \bibinfo{person}{Mohan~M. Trivedi}.} \bibinfo{year}{2018}\natexlab{a}.
\newblock \showarticletitle{Convolutional Social Pooling for Vehicle Trajectory Prediction}. In \bibinfo{booktitle}{\emph{CVPR}}. \bibinfo{pages}{1468--1476}.
\newblock


\bibitem[Deo and Trivedi(2018b)]%
        {DBLP:conf/ivs/DeoT18}
\bibfield{author}{\bibinfo{person}{Nachiket Deo} {and} \bibinfo{person}{Mohan~M. Trivedi}.} \bibinfo{year}{2018}\natexlab{b}.
\newblock \showarticletitle{Multi-Modal Trajectory Prediction of Surrounding Vehicles with Maneuver based LSTMs}. In \bibinfo{booktitle}{\emph{IV}}. \bibinfo{pages}{1179--1184}.
\newblock


\bibitem[Durrant and McCammon(2011)]%
        {durrant2011molecular}
\bibfield{author}{\bibinfo{person}{Jacob~D Durrant} {and} \bibinfo{person}{J~Andrew McCammon}.} \bibinfo{year}{2011}\natexlab{}.
\newblock \showarticletitle{Molecular dynamics simulations and drug discovery}.
\newblock \bibinfo{journal}{\emph{BMC biology}} \bibinfo{volume}{9}, \bibinfo{number}{1} (\bibinfo{year}{2011}), \bibinfo{pages}{1--9}.
\newblock


\bibitem[Feng et~al\mbox{.}(2024)]%
        {feng2024deep}
\bibfield{author}{\bibinfo{person}{Tao Feng}, \bibinfo{person}{Ziqi Gao}, \bibinfo{person}{Jiaxuan You}, \bibinfo{person}{Chenyi Zi}, \bibinfo{person}{Yan Zhou}, \bibinfo{person}{Chen Zhang}, {and} \bibinfo{person}{Jia Li}.} \bibinfo{year}{2024}\natexlab{}.
\newblock \showarticletitle{Deep Reinforcement Learning for Modelling Protein Complexes}.
\newblock \bibinfo{journal}{\emph{arXiv preprint arXiv:2405.02299}} (\bibinfo{year}{2024}).
\newblock


\bibitem[Fey and Lenssen(2019)]%
        {fey2019fast}
\bibfield{author}{\bibinfo{person}{Matthias Fey} {and} \bibinfo{person}{Jan~Eric Lenssen}.} \bibinfo{year}{2019}\natexlab{}.
\newblock \showarticletitle{Fast graph representation learning with PyTorch Geometric}.
\newblock \bibinfo{journal}{\emph{arXiv preprint arXiv:1903.02428}} (\bibinfo{year}{2019}).
\newblock


\bibitem[Finzi et~al\mbox{.}(2020)]%
        {finzi2020generalizing}
\bibfield{author}{\bibinfo{person}{Marc Finzi}, \bibinfo{person}{Samuel Stanton}, \bibinfo{person}{Pavel Izmailov}, {and} \bibinfo{person}{Andrew~Gordon Wilson}.} \bibinfo{year}{2020}\natexlab{}.
\newblock \showarticletitle{Generalizing convolutional neural networks for equivariance to lie groups on arbitrary continuous data}. In \bibinfo{booktitle}{\emph{ICML}}. PMLR, \bibinfo{pages}{3165--3176}.
\newblock


\bibitem[Fiorini and Shiller(1998)]%
        {fiorini1998motion}
\bibfield{author}{\bibinfo{person}{Paolo Fiorini} {and} \bibinfo{person}{Zvi Shiller}.} \bibinfo{year}{1998}\natexlab{}.
\newblock \showarticletitle{Motion planning in dynamic environments using velocity obstacles}.
\newblock \bibinfo{journal}{\emph{IJRR}} \bibinfo{volume}{17}, \bibinfo{number}{7} (\bibinfo{year}{1998}), \bibinfo{pages}{760--772}.
\newblock


\bibitem[Fuchs et~al\mbox{.}(2020)]%
        {fuchs2020se}
\bibfield{author}{\bibinfo{person}{Fabian Fuchs}, \bibinfo{person}{Daniel Worrall}, \bibinfo{person}{Volker Fischer}, {and} \bibinfo{person}{Max Welling}.} \bibinfo{year}{2020}\natexlab{}.
\newblock \showarticletitle{Se (3)-transformers: 3d roto-translation equivariant attention networks}.
\newblock \bibinfo{journal}{\emph{NeurIPS}} (\bibinfo{year}{2020}).
\newblock


\bibitem[Gao et~al\mbox{.}(2023)]%
        {gao2023hierarchical}
\bibfield{author}{\bibinfo{person}{Ziqi Gao}, \bibinfo{person}{Chenran Jiang}, \bibinfo{person}{Jiawen Zhang}, \bibinfo{person}{Xiaosen Jiang}, \bibinfo{person}{Lanqing Li}, \bibinfo{person}{Peilin Zhao}, \bibinfo{person}{Huanming Yang}, \bibinfo{person}{Yong Huang}, {and} \bibinfo{person}{Jia Li}.} \bibinfo{year}{2023}\natexlab{}.
\newblock \showarticletitle{Hierarchical graph learning for protein--protein interaction}.
\newblock \bibinfo{journal}{\emph{Nature Communications}} \bibinfo{volume}{14}, \bibinfo{number}{1} (\bibinfo{year}{2023}), \bibinfo{pages}{1093}.
\newblock


\bibitem[Giuliari et~al\mbox{.}(2021)]%
        {giuliari2021transformer}
\bibfield{author}{\bibinfo{person}{Francesco Giuliari}, \bibinfo{person}{Irtiza Hasan}, \bibinfo{person}{Marco Cristani}, {and} \bibinfo{person}{Fabio Galasso}.} \bibinfo{year}{2021}\natexlab{}.
\newblock \showarticletitle{Transformer networks for trajectory forecasting}. In \bibinfo{booktitle}{\emph{ICPR}}. IEEE, \bibinfo{pages}{10335--10342}.
\newblock


\bibitem[Guo et~al\mbox{.}(2017)]%
        {DBLP:conf/ijcai/GuoTYLH17}
\bibfield{author}{\bibinfo{person}{Huifeng Guo}, \bibinfo{person}{Ruiming Tang}, \bibinfo{person}{Yunming Ye}, \bibinfo{person}{Zhenguo Li}, {and} \bibinfo{person}{Xiuqiang He}.} \bibinfo{year}{2017}\natexlab{}.
\newblock \showarticletitle{DeepFM: {A} Factorization-Machine based Neural Network for {CTR} Prediction}. In \bibinfo{booktitle}{\emph{IJCAI}}. \bibinfo{pages}{1725--1731}.
\newblock


\bibitem[Han et~al\mbox{.}(2024)]%
        {han2024survey}
\bibfield{author}{\bibinfo{person}{Jiaqi Han}, \bibinfo{person}{Jiacheng Cen}, \bibinfo{person}{Liming Wu}, \bibinfo{person}{Zongzhao Li}, \bibinfo{person}{Xiangzhe Kong}, \bibinfo{person}{Rui Jiao}, \bibinfo{person}{Ziyang Yu}, \bibinfo{person}{Tingyang Xu}, \bibinfo{person}{Fandi Wu}, \bibinfo{person}{Zihe Wang}, {et~al\mbox{.}}} \bibinfo{year}{2024}\natexlab{}.
\newblock \showarticletitle{A Survey of Geometric Graph Neural Networks: Data Structures, Models and Applications}.
\newblock \bibinfo{journal}{\emph{arXiv preprint arXiv:2403.00485}} (\bibinfo{year}{2024}).
\newblock


\bibitem[Han et~al\mbox{.}(2022a)]%
        {DBLP:conf/nips/Han0XR22}
\bibfield{author}{\bibinfo{person}{Jiaqi Han}, \bibinfo{person}{Wenbing Huang}, \bibinfo{person}{Tingyang Xu}, {and} \bibinfo{person}{Yu Rong}.} \bibinfo{year}{2022}\natexlab{a}.
\newblock \showarticletitle{Equivariant Graph Hierarchy-Based Neural Networks}. In \bibinfo{booktitle}{\emph{NeurIPS}}.
\newblock


\bibitem[Han et~al\mbox{.}(2022b)]%
        {DBLP:journals/corr/abs-2202-07230}
\bibfield{author}{\bibinfo{person}{Jiaqi Han}, \bibinfo{person}{Yu Rong}, \bibinfo{person}{Tingyang Xu}, {and} \bibinfo{person}{Wenbing Huang}.} \bibinfo{year}{2022}\natexlab{b}.
\newblock \showarticletitle{Geometrically Equivariant Graph Neural Networks: {A} Survey}.
\newblock \bibinfo{journal}{\emph{CoRR}}  \bibinfo{volume}{abs/2202.07230} (\bibinfo{year}{2022}).
\newblock


\bibitem[Helbing and Molnar(1995)]%
        {helbing1995social}
\bibfield{author}{\bibinfo{person}{Dirk Helbing} {and} \bibinfo{person}{Peter Molnar}.} \bibinfo{year}{1995}\natexlab{}.
\newblock \showarticletitle{Social force model for pedestrian dynamics}.
\newblock \bibinfo{journal}{\emph{Physical review E}} \bibinfo{volume}{51}, \bibinfo{number}{5} (\bibinfo{year}{1995}), \bibinfo{pages}{4282}.
\newblock


\bibitem[Huang et~al\mbox{.}(2022)]%
        {DBLP:conf/iclr/0001HRX0H22}
\bibfield{author}{\bibinfo{person}{Wenbing Huang}, \bibinfo{person}{Jiaqi Han}, \bibinfo{person}{Yu Rong}, \bibinfo{person}{Tingyang Xu}, \bibinfo{person}{Fuchun Sun}, {and} \bibinfo{person}{Junzhou Huang}.} \bibinfo{year}{2022}\natexlab{}.
\newblock \showarticletitle{Equivariant Graph Mechanics Networks with Constraints}. In \bibinfo{booktitle}{\emph{ICLR}}.
\newblock


\bibitem[Huang et~al\mbox{.}(2023)]%
        {huang2023generalizing}
\bibfield{author}{\bibinfo{person}{Zijie Huang}, \bibinfo{person}{Yizhou Sun}, {and} \bibinfo{person}{Wei Wang}.} \bibinfo{year}{2023}\natexlab{}.
\newblock \showarticletitle{Generalizing graph ode for learning complex system dynamics across environments}. In \bibinfo{booktitle}{\emph{Proceedings of the 29th ACM SIGKDD Conference on Knowledge Discovery and Data Mining}}. \bibinfo{pages}{798--809}.
\newblock


\bibitem[Hutchinson et~al\mbox{.}(2021)]%
        {hutchinson2021lietransformer}
\bibfield{author}{\bibinfo{person}{Michael~J Hutchinson}, \bibinfo{person}{Charline Le~Lan}, \bibinfo{person}{Sheheryar Zaidi}, \bibinfo{person}{Emilien Dupont}, \bibinfo{person}{Yee~Whye Teh}, {and} \bibinfo{person}{Hyunjik Kim}.} \bibinfo{year}{2021}\natexlab{}.
\newblock \showarticletitle{Lietransformer: equivariant self-attention for lie groups}. In \bibinfo{booktitle}{\emph{ICML}}. PMLR, \bibinfo{pages}{4533--4543}.
\newblock


\bibitem[Johansson et~al\mbox{.}(2008)]%
        {johansson2008crowd}
\bibfield{author}{\bibinfo{person}{Anders Johansson}, \bibinfo{person}{Dirk Helbing}, \bibinfo{person}{Habib~Z Al-Abideen}, {and} \bibinfo{person}{Salim Al-Bosta}.} \bibinfo{year}{2008}\natexlab{}.
\newblock \showarticletitle{From crowd dynamics to crowd safety: a video-based analysis}.
\newblock \bibinfo{journal}{\emph{Advances in Complex Systems}} \bibinfo{volume}{11}, \bibinfo{number}{04} (\bibinfo{year}{2008}), \bibinfo{pages}{497--527}.
\newblock


\bibitem[Kaba and Ravanbakhsh(2022)]%
        {DBLP:conf/nips/KabaR22}
\bibfield{author}{\bibinfo{person}{S{\'{e}}kou{-}Oumar Kaba} {and} \bibinfo{person}{Siamak Ravanbakhsh}.} \bibinfo{year}{2022}\natexlab{}.
\newblock \showarticletitle{Equivariant Networks for Crystal Structures}. In \bibinfo{booktitle}{\emph{NeurIPS}}.
\newblock


\bibitem[Kipf et~al\mbox{.}(2018)]%
        {kipf2018neural}
\bibfield{author}{\bibinfo{person}{Thomas Kipf}, \bibinfo{person}{Ethan Fetaya}, \bibinfo{person}{Kuan-Chieh Wang}, \bibinfo{person}{Max Welling}, {and} \bibinfo{person}{Richard Zemel}.} \bibinfo{year}{2018}\natexlab{}.
\newblock \showarticletitle{Neural relational inference for interacting systems}. In \bibinfo{booktitle}{\emph{International conference on machine learning}}. PMLR, \bibinfo{pages}{2688--2697}.
\newblock


\bibitem[K{\"o}hler et~al\mbox{.}(2019)]%
        {kohler2019equivariant}
\bibfield{author}{\bibinfo{person}{Jonas K{\"o}hler}, \bibinfo{person}{Leon Klein}, {and} \bibinfo{person}{Frank No{\'e}}.} \bibinfo{year}{2019}\natexlab{}.
\newblock \showarticletitle{Equivariant flows: sampling configurations for multi-body systems with symmetric energies}.
\newblock \bibinfo{journal}{\emph{arXiv preprint arXiv:1910.00753}} (\bibinfo{year}{2019}).
\newblock


\bibitem[Kothari et~al\mbox{.}(2022)]%
        {DBLP:journals/tits/KothariKA22}
\bibfield{author}{\bibinfo{person}{Parth Kothari}, \bibinfo{person}{Sven Kreiss}, {and} \bibinfo{person}{Alexandre Alahi}.} \bibinfo{year}{2022}\natexlab{}.
\newblock \showarticletitle{Human Trajectory Forecasting in Crowds: {A} Deep Learning Perspective}.
\newblock \bibinfo{journal}{\emph{{IEEE} Trans. Intell. Transp. Syst.}} \bibinfo{volume}{23}, \bibinfo{number}{7} (\bibinfo{year}{2022}), \bibinfo{pages}{7386--7400}.
\newblock


\bibitem[Krajewski et~al\mbox{.}(2018)]%
        {highDdataset}
\bibfield{author}{\bibinfo{person}{Robert Krajewski}, \bibinfo{person}{Julian Bock}, \bibinfo{person}{Laurent Kloeker}, {and} \bibinfo{person}{Lutz Eckstein}.} \bibinfo{year}{2018}\natexlab{}.
\newblock \showarticletitle{The highD Dataset: A Drone Dataset of Naturalistic Vehicle Trajectories on German Highways for Validation of Highly Automated Driving Systems}. In \bibinfo{booktitle}{\emph{ITSC}}.
\newblock


\bibitem[Li et~al\mbox{.}(2023)]%
        {li2023survey}
\bibfield{author}{\bibinfo{person}{Yuhan Li}, \bibinfo{person}{Zhixun Li}, \bibinfo{person}{Peisong Wang}, \bibinfo{person}{Jia Li}, \bibinfo{person}{Xiangguo Sun}, \bibinfo{person}{Hong Cheng}, {and} \bibinfo{person}{Jeffrey~Xu Yu}.} \bibinfo{year}{2023}\natexlab{}.
\newblock \showarticletitle{A survey of graph meets large language model: Progress and future directions}.
\newblock \bibinfo{journal}{\emph{arXiv preprint arXiv:2311.12399}} (\bibinfo{year}{2023}).
\newblock


\bibitem[Li and Farimani(2022)]%
        {li2022graph}
\bibfield{author}{\bibinfo{person}{Zijie Li} {and} \bibinfo{person}{Amir~Barati Farimani}.} \bibinfo{year}{2022}\natexlab{}.
\newblock \showarticletitle{Graph neural network-accelerated Lagrangian fluid simulation}.
\newblock \bibinfo{journal}{\emph{Computers \& Graphics}}  \bibinfo{volume}{103} (\bibinfo{year}{2022}), \bibinfo{pages}{201--211}.
\newblock


\bibitem[Liu et~al\mbox{.}(2020)]%
        {liu2020modelling}
\bibfield{author}{\bibinfo{person}{Yang Liu}, \bibinfo{person}{Liang Chen}, \bibinfo{person}{Xiangnan He}, \bibinfo{person}{Jiaying Peng}, \bibinfo{person}{Zibin Zheng}, {and} \bibinfo{person}{Jie Tang}.} \bibinfo{year}{2020}\natexlab{}.
\newblock \showarticletitle{Modelling high-order social relations for item recommendation}.
\newblock \bibinfo{journal}{\emph{IEEE Transactions on Knowledge and Data Engineering}} \bibinfo{volume}{34}, \bibinfo{number}{9} (\bibinfo{year}{2020}), \bibinfo{pages}{4385--4397}.
\newblock


\bibitem[Liu et~al\mbox{.}(2023a)]%
        {liu2023improving}
\bibfield{author}{\bibinfo{person}{Yang Liu}, \bibinfo{person}{Jiashun Cheng}, \bibinfo{person}{Haihong Zhao}, \bibinfo{person}{Tingyang Xu}, \bibinfo{person}{Peilin Zhao}, \bibinfo{person}{Fugee Tsung}, \bibinfo{person}{Jia Li}, {and} \bibinfo{person}{Yu Rong}.} \bibinfo{year}{2023}\natexlab{a}.
\newblock \showarticletitle{SEGNO: Generalizing Equivariant Graph Neural Networks with Physical Inductive Biases}. In \bibinfo{booktitle}{\emph{ICLR}}.
\newblock


\bibitem[Liu et~al\mbox{.}(2023b)]%
        {liu2023human}
\bibfield{author}{\bibinfo{person}{Yang Liu}, \bibinfo{person}{Yu Rong}, \bibinfo{person}{Zhuoning Guo}, \bibinfo{person}{Nuo Chen}, \bibinfo{person}{Tingyang Xu}, \bibinfo{person}{Fugee Tsung}, {and} \bibinfo{person}{Jia Li}.} \bibinfo{year}{2023}\natexlab{b}.
\newblock \showarticletitle{Human mobility modeling during the COVID-19 pandemic via deep graph diffusion infomax}. In \bibinfo{booktitle}{\emph{AAAI}}, Vol.~\bibinfo{volume}{37}. \bibinfo{pages}{14347--14355}.
\newblock


\bibitem[Ma et~al\mbox{.}(2022)]%
        {ma2022cross}
\bibfield{author}{\bibinfo{person}{Hehuan Ma}, \bibinfo{person}{Yatao Bian}, \bibinfo{person}{Yu Rong}, \bibinfo{person}{Wenbing Huang}, \bibinfo{person}{Tingyang Xu}, \bibinfo{person}{Weiyang Xie}, \bibinfo{person}{Geyan Ye}, {and} \bibinfo{person}{Junzhou Huang}.} \bibinfo{year}{2022}\natexlab{}.
\newblock \showarticletitle{Cross-dependent graph neural networks for molecular property prediction}.
\newblock \bibinfo{journal}{\emph{Bioinformatics}} \bibinfo{volume}{38}, \bibinfo{number}{7} (\bibinfo{year}{2022}), \bibinfo{pages}{2003--2009}.
\newblock


\bibitem[Mrowca et~al\mbox{.}(2018)]%
        {mrowca2018flexible}
\bibfield{author}{\bibinfo{person}{Damian Mrowca}, \bibinfo{person}{Chengxu Zhuang}, \bibinfo{person}{Elias Wang}, \bibinfo{person}{Nick Haber}, \bibinfo{person}{Li~F Fei-Fei}, \bibinfo{person}{Josh Tenenbaum}, {and} \bibinfo{person}{Daniel~L Yamins}.} \bibinfo{year}{2018}\natexlab{}.
\newblock \showarticletitle{Flexible neural representation for physics prediction}.
\newblock \bibinfo{journal}{\emph{Advances in neural information processing systems}}  \bibinfo{volume}{31} (\bibinfo{year}{2018}).
\newblock


\bibitem[Reynolds(1987)]%
        {reynolds1987flocks}
\bibfield{author}{\bibinfo{person}{Craig~W Reynolds}.} \bibinfo{year}{1987}\natexlab{}.
\newblock \showarticletitle{Flocks, herds and schools: A distributed behavioral model}. In \bibinfo{booktitle}{\emph{SIGGRAPH}}. \bibinfo{pages}{25--34}.
\newblock


\bibitem[Saboia and Goldenstein(2012)]%
        {saboia2012crowd}
\bibfield{author}{\bibinfo{person}{Priscila Saboia} {and} \bibinfo{person}{Siome Goldenstein}.} \bibinfo{year}{2012}\natexlab{}.
\newblock \showarticletitle{Crowd simulation: applying mobile grids to the social force model}.
\newblock \bibinfo{journal}{\emph{The Visual Computer}}  \bibinfo{volume}{28} (\bibinfo{year}{2012}), \bibinfo{pages}{1039--1048}.
\newblock


\bibitem[Salzmann et~al\mbox{.}(2020)]%
        {DBLP:conf/eccv/SalzmannICP20}
\bibfield{author}{\bibinfo{person}{Tim Salzmann}, \bibinfo{person}{Boris Ivanovic}, \bibinfo{person}{Punarjay Chakravarty}, {and} \bibinfo{person}{Marco Pavone}.} \bibinfo{year}{2020}\natexlab{}.
\newblock \showarticletitle{Trajectron++: Dynamically-Feasible Trajectory Forecasting with Heterogeneous Data}. In \bibinfo{booktitle}{\emph{ECCV}}, Vol.~\bibinfo{volume}{12363}. \bibinfo{pages}{683--700}.
\newblock


\bibitem[Sanchez{-}Gonzalez et~al\mbox{.}(2020)]%
        {DBLP:conf/icml/Sanchez-Gonzalez20}
\bibfield{author}{\bibinfo{person}{Alvaro Sanchez{-}Gonzalez}, \bibinfo{person}{Jonathan Godwin}, \bibinfo{person}{Tobias Pfaff}, \bibinfo{person}{Rex Ying}, \bibinfo{person}{Jure Leskovec}, {and} \bibinfo{person}{Peter~W. Battaglia}.} \bibinfo{year}{2020}\natexlab{}.
\newblock \showarticletitle{Learning to Simulate Complex Physics with Graph Networks}. In \bibinfo{booktitle}{\emph{ICML}}, Vol.~\bibinfo{volume}{119}. \bibinfo{pages}{8459--8468}.
\newblock


\bibitem[Satorras et~al\mbox{.}(2021)]%
        {satorras2021n}
\bibfield{author}{\bibinfo{person}{V{\i}ctor~Garcia Satorras}, \bibinfo{person}{Emiel Hoogeboom}, {and} \bibinfo{person}{Max Welling}.} \bibinfo{year}{2021}\natexlab{}.
\newblock \showarticletitle{E (n) equivariant graph neural networks}. In \bibinfo{booktitle}{\emph{ICML}}. \bibinfo{pages}{9323--9332}.
\newblock


\bibitem[Shi et~al\mbox{.}(2023)]%
        {DBLP:conf/www/ShiYL23}
\bibfield{author}{\bibinfo{person}{Hongzhi Shi}, \bibinfo{person}{Quanming Yao}, {and} \bibinfo{person}{Yong Li}.} \bibinfo{year}{2023}\natexlab{}.
\newblock \showarticletitle{Learning to Simulate Crowd Trajectories with Graph Networks}. In \bibinfo{booktitle}{\emph{WWW}}. \bibinfo{pages}{4200--4209}.
\newblock


\bibitem[Suzuki(1986)]%
        {suzuki1986group}
\bibfield{author}{\bibinfo{person}{Michio Suzuki}.} \bibinfo{year}{1986}\natexlab{}.
\newblock \bibinfo{booktitle}{\emph{Group theory II}}.
\newblock \bibinfo{publisher}{Springer}.
\newblock


\bibitem[Thomas et~al\mbox{.}(2018)]%
        {thomas2018tensor}
\bibfield{author}{\bibinfo{person}{Nathaniel Thomas}, \bibinfo{person}{Tess Smidt}, \bibinfo{person}{Steven Kearnes}, \bibinfo{person}{Lusann Yang}, \bibinfo{person}{Li Li}, \bibinfo{person}{Kai Kohlhoff}, {and} \bibinfo{person}{Patrick Riley}.} \bibinfo{year}{2018}\natexlab{}.
\newblock \showarticletitle{Tensor field networks: Rotation-and translation-equivariant neural networks for 3d point clouds}.
\newblock \bibinfo{journal}{\emph{arXiv:1802.08219}} (\bibinfo{year}{2018}).
\newblock


\bibitem[Unke et~al\mbox{.}(2021)]%
        {unke2021machine}
\bibfield{author}{\bibinfo{person}{Oliver~T Unke}, \bibinfo{person}{Stefan Chmiela}, \bibinfo{person}{Huziel~E Sauceda}, \bibinfo{person}{Michael Gastegger}, \bibinfo{person}{Igor Poltavskyi}, \bibinfo{person}{Kristof~T Sch{\"u}tt}, \bibinfo{person}{Alexandre Tkatchenko}, {and} \bibinfo{person}{Klaus-Robert M{\"u}ller}.} \bibinfo{year}{2021}\natexlab{}.
\newblock \showarticletitle{Machine Learning Force Fields}.
\newblock \bibinfo{journal}{\emph{Chemical Reviews}} (\bibinfo{year}{2021}).
\newblock


\bibitem[Walters et~al\mbox{.}(2021)]%
        {DBLP:conf/iclr/WaltersLY21}
\bibfield{author}{\bibinfo{person}{Robin Walters}, \bibinfo{person}{Jinxi Li}, {and} \bibinfo{person}{Rose Yu}.} \bibinfo{year}{2021}\natexlab{}.
\newblock \showarticletitle{Trajectory Prediction using Equivariant Continuous Convolution}. In \bibinfo{booktitle}{\emph{ICLR}}. \bibinfo{publisher}{OpenReview.net}.
\newblock


\bibitem[Wang and Chodera(2023)]%
        {DBLP:conf/iclr/WangC23}
\bibfield{author}{\bibinfo{person}{Yuanqing Wang} {and} \bibinfo{person}{John~D. Chodera}.} \bibinfo{year}{2023}\natexlab{}.
\newblock \showarticletitle{Spatial Attention Kinetic Networks with E(n)-Equivariance}. In \bibinfo{booktitle}{\emph{ICLR}}. \bibinfo{publisher}{OpenReview.net}.
\newblock


\bibitem[Wu et~al\mbox{.}(2024)]%
        {wu2024equivariant}
\bibfield{author}{\bibinfo{person}{Liming Wu}, \bibinfo{person}{Zhichao Hou}, \bibinfo{person}{Jirui Yuan}, \bibinfo{person}{Yu Rong}, {and} \bibinfo{person}{Wenbing Huang}.} \bibinfo{year}{2024}\natexlab{}.
\newblock \showarticletitle{Equivariant Spatio-Temporal Attentive Graph Networks to Simulate Physical Dynamics}.
\newblock \bibinfo{journal}{\emph{Advances in Neural Information Processing Systems}}  \bibinfo{volume}{36} (\bibinfo{year}{2024}).
\newblock


\bibitem[Xu et~al\mbox{.}(2023)]%
        {xu2023eqmotion}
\bibfield{author}{\bibinfo{person}{Chenxin Xu}, \bibinfo{person}{Robby~T Tan}, \bibinfo{person}{Yuhong Tan}, \bibinfo{person}{Siheng Chen}, \bibinfo{person}{Yu~Guang Wang}, \bibinfo{person}{Xinchao Wang}, {and} \bibinfo{person}{Yanfeng Wang}.} \bibinfo{year}{2023}\natexlab{}.
\newblock \showarticletitle{EqMotion: Equivariant Multi-agent Motion Prediction with Invariant Interaction Reasoning}. In \bibinfo{booktitle}{\emph{CVPR}}. \bibinfo{pages}{1410--1420}.
\newblock


\bibitem[Yang et~al\mbox{.}(2021)]%
        {yang2021hierarchical}
\bibfield{author}{\bibinfo{person}{Jinyu Yang}, \bibinfo{person}{Peilin Zhao}, \bibinfo{person}{Yu Rong}, \bibinfo{person}{Chaochao Yan}, \bibinfo{person}{Chunyuan Li}, \bibinfo{person}{Hehuan Ma}, {and} \bibinfo{person}{Junzhou Huang}.} \bibinfo{year}{2021}\natexlab{}.
\newblock \showarticletitle{Hierarchical graph capsule network}. In \bibinfo{booktitle}{\emph{AAAI}}.
\newblock


\bibitem[Yang et~al\mbox{.}(2020)]%
        {DBLP:journals/cvgip/YangLGPH20}
\bibfield{author}{\bibinfo{person}{Shanwen Yang}, \bibinfo{person}{Tianrui Li}, \bibinfo{person}{Xun Gong}, \bibinfo{person}{Bo Peng}, {and} \bibinfo{person}{Jie Hu}.} \bibinfo{year}{2020}\natexlab{}.
\newblock \showarticletitle{A review on crowd simulation and modeling}.
\newblock \bibinfo{journal}{\emph{Graph. Model.}}  \bibinfo{volume}{111} (\bibinfo{year}{2020}), \bibinfo{pages}{101081}.
\newblock


\bibitem[Yu et~al\mbox{.}(2023)]%
        {DBLP:conf/cikm/YuZLJ23}
\bibfield{author}{\bibinfo{person}{Zihan Yu}, \bibinfo{person}{Guozhen Zhang}, \bibinfo{person}{Yong Li}, {and} \bibinfo{person}{Depeng Jin}.} \bibinfo{year}{2023}\natexlab{}.
\newblock \showarticletitle{Understanding and Modeling Collision Avoidance Behavior for Realistic Crowd Simulation}. In \bibinfo{booktitle}{\emph{CIKM}}.
\newblock


\bibitem[Yuan et~al\mbox{.}(2021)]%
        {DBLP:conf/iccv/0007WOK21}
\bibfield{author}{\bibinfo{person}{Ye Yuan}, \bibinfo{person}{Xinshuo Weng}, \bibinfo{person}{Yanglan Ou}, {and} \bibinfo{person}{Kris Kitani}.} \bibinfo{year}{2021}\natexlab{}.
\newblock \showarticletitle{AgentFormer: Agent-Aware Transformers for Socio-Temporal Multi-Agent Forecasting}. In \bibinfo{booktitle}{\emph{ICCV}}.
\newblock


\bibitem[Zhao et~al\mbox{.}(2024a)]%
        {zhao2024all}
\bibfield{author}{\bibinfo{person}{Haihong Zhao}, \bibinfo{person}{Aochuan Chen}, \bibinfo{person}{Xiangguo Sun}, \bibinfo{person}{Hong Cheng}, {and} \bibinfo{person}{Jia Li}.} \bibinfo{year}{2024}\natexlab{a}.
\newblock \showarticletitle{All in One and One for All: A Simple yet Effective Method towards Cross-domain Graph Pretraining}.
\newblock \bibinfo{journal}{\emph{arXiv preprint arXiv:2402.09834}} (\bibinfo{year}{2024}).
\newblock


\bibitem[Zhao et~al\mbox{.}(2023)]%
        {zhao2023effective}
\bibfield{author}{\bibinfo{person}{Haihong Zhao}, \bibinfo{person}{Bo Yang}, \bibinfo{person}{Jiaxu Cui}, \bibinfo{person}{Qianli Xing}, \bibinfo{person}{Jiaxing Shen}, \bibinfo{person}{Fujin Zhu}, {and} \bibinfo{person}{Jiannong Cao}.} \bibinfo{year}{2023}\natexlab{}.
\newblock \showarticletitle{Effective fault scenario identification for communication networks via knowledge-enhanced graph neural networks}.
\newblock \bibinfo{journal}{\emph{IEEE Transactions on Mobile Computing}} (\bibinfo{year}{2023}).
\newblock


\bibitem[Zhao et~al\mbox{.}(2024b)]%
        {zhao2024weakly}
\bibfield{author}{\bibinfo{person}{Haihong Zhao}, \bibinfo{person}{Chenyi Zi}, \bibinfo{person}{Yang Liu}, \bibinfo{person}{Chen Zhang}, \bibinfo{person}{Yan Zhou}, {and} \bibinfo{person}{Jia Li}.} \bibinfo{year}{2024}\natexlab{b}.
\newblock \showarticletitle{Weakly Supervised Anomaly Detection via Knowledge-Data Alignment}. In \bibinfo{booktitle}{\emph{Proceedings of the ACM on Web Conference 2024}}. \bibinfo{pages}{4083--4094}.
\newblock


\end{thebibliography}

\appendix

\section{Proof}

\subsection{Proof of Theorem~\ref{thm:equ}}\label{proof:thm}
Since the set is unordered and the point group satisfied closure, let $\bm{O}_1, \bm{O}_2\in P$ be transformations in point group $P$, we have $\bm{O}_1\bm{O}_2\bm{x}\in \{\bm{O}\bm{x}\}_{\bm{O}\in P}$. Given that $\phi$ is a permutation-invariant function, for any transformation $\bm{O}_1\in P$, we have
\vspace{-1ex}
\begin{equation}
\vspace{-1ex}
\begin{aligned}
    \mu_P(\bm{O}_1\bm{x}, \bm{h}) &= (\bm{O}_1\bm{q}_i - \bm{O}_1\bm{q}_j)\phi(\{\bm{O}_2\bm{O}_1\bm{x}\}_{\bm{O}_1,\bm{O}_2\in P}, \bm{h})\\
    &= \bm{O}_1(\bm{q}_i - \bm{q}_j)\phi(\{\bm{O}\bm{x}\}_{\bm{O}\in P}, \bm{h}) = \bm{O}_1\mu_P(\bm{x}, \bm{h})
\end{aligned}
\end{equation}
Therefore, $\mu_P$ is equivariant to the point group $P$ if $\phi$ is permutation-invariant.

\subsection{Proof of Proposition~\ref{thm:prop}}\label{proof:prop}
For simplicity and with no loss of generality, the message embedding in Eq.~\ref{eq:attn} can be rewritten as 
\begin{equation}
\begin{aligned}
\bm{m} &= \sum_{j=1}^{d}\frac{\text{exp}(<\sigma_q (\bm{e}_{i}), \sigma_k (\bm{e}_{j})>)\sigma_v(\bm{e}_{j})}{\sum_{j=1}^{d}\text{exp}(<\sigma_q (\bm{e}_{i}), \sigma_k (\bm{e}_{j})>)}.
\end{aligned}
\end{equation}
Then the message embedding $\bm{m}'$ for any permutation $\pi\in\Pi_{d}$ is:
\begin{equation}
\begin{aligned}
\bm{m}' &= \sum_{j=1}^{d}\frac{\text{exp}(<\sigma_q (\bm{e}_{\pi(i)}), \sigma_k (\bm{e}_{\pi(j)})>)\sigma_v(\bm{e}_{\pi(j)})}{\sum_{j=1}^{d}\text{exp}(<\sigma_q (\bm{e}_{\pi(i)}), \sigma_k (\bm{e}_{\pi(j)})>)}\\
&= \sum_{j=1}^{d}\frac{\text{exp}(<\sigma_q (\bm{e}_{\pi(i)}), \sigma_k (\bm{e}_{\pi(j)})>)\sigma_v(\bm{e}_{\pi(j)})}{\sum_{j=1}^{d}\text{exp}(<\sigma_q (\bm{e}_{\pi(i)}), \sigma_k (\bm{e}_{j})>)}\\
&= \sum_{j=1}^{d}\frac{\text{exp}(<\sigma_q (\bm{e}_{\pi(i)}), \sigma_k (\bm{e}_{j})>)\sigma_v(\bm{e}_{j})}{\sum_{j=1}^{d}\text{exp}(<\sigma_q (\bm{e}_{\pi(i)}), \sigma_k (\bm{e}_{j})>)}.
\end{aligned}
\end{equation}
The second and third equation hold because sum is permutation-invariant. Finally, since $\bm{e}_{\pi(i)}=\bm{e}_{i}$, $\bm{m}' = \bm{m}$.

\begin{figure}[t]
    \centering
    \includegraphics[width=0.5\textwidth]{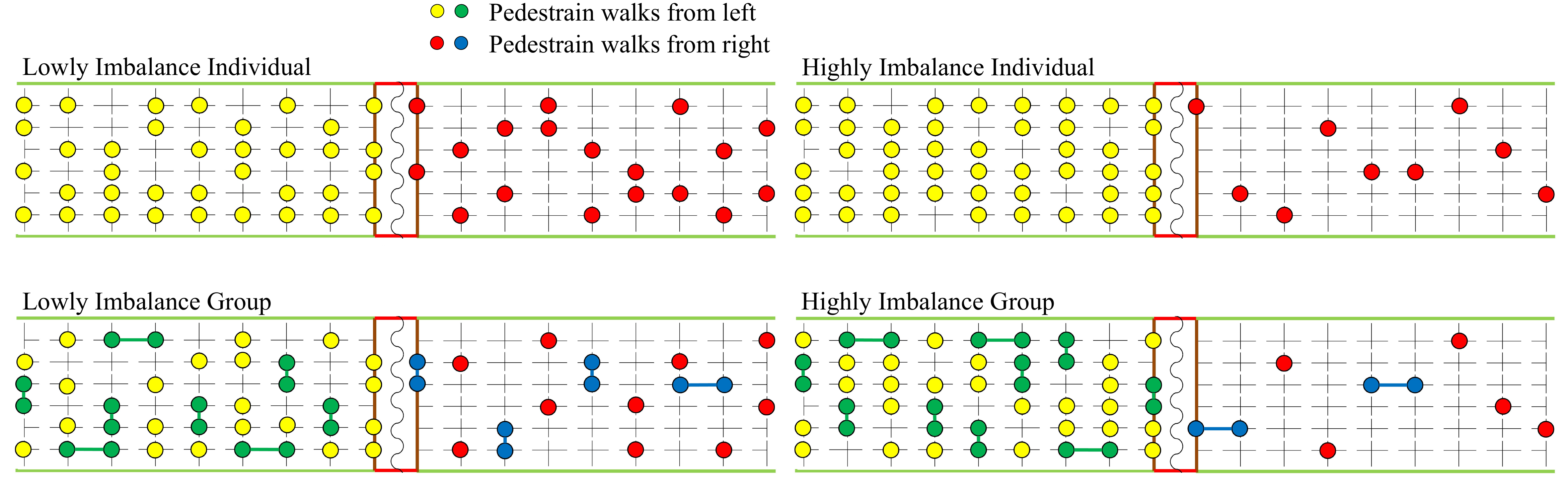}
    \vspace{-2ex}
    \caption{Illustrations of starting positions of crowd scenes.}
        \vspace{-2ex}
    \label{fig:crowd_img}
\end{figure}

\begin{figure}[t]
    \centering
    \includegraphics[width=0.5\textwidth]{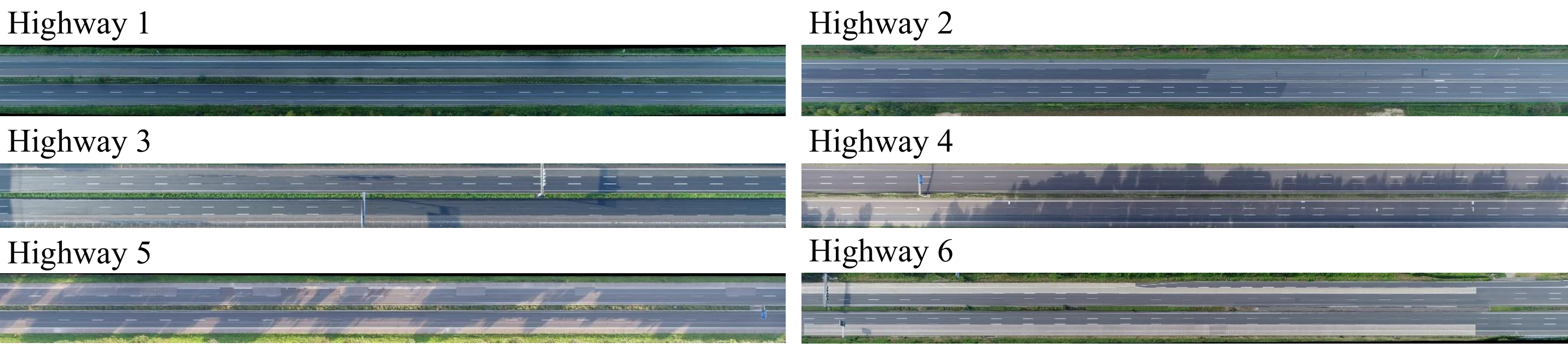}
    \caption{Representative highway scene.}
    \label{fig:highway_img}
\end{figure}

\section{More Details on Datasets}\label{sec:details}
\subsubsection{\textbf{Charged N-body system}}
We adopt the 3D N-body simulation code introduced by \cite{satorras2021n}, which builds upon the 2D implementation from \cite{kipf2018neural} and extends it to three dimensions. System trajectories are generated at a 0.02 time step and are constrained with four different sizes of virtual boxes: 5 × 5 × 5, 5 × 4 × 4, 5 × 4 × 3, and 4 × 4 × 4, respectively. Initial locations are sampled from a Gaussian distribution (mean $\mu=0$, standard deviation $\sigma=1$), and the initial velocity is a random vector with a norm of 0.5.
\subsubsection{\textbf{Gravity N-body system}}
The gravitational N-body systems code is sourced from ~\cite{DBLP:conf/iclr/BrandstetterHPB22}, implemented within the same framework as the aforementioned charged N-body systems. System trajectories are generated at a time step of 0.02, constrained within four different sizes of virtual boxes 5 × 5 × 5, 5 × 4 × 4, 5 × 4 × 3, and 4 × 4 × 4, respectively. Initial locations are sampled from a Gaussian distribution (mean $\mu=0$, standard deviation $\sigma=1$), and the initial velocity is a random vector with a norm of 0.5. Particle mass is consistently set to one. 

\quad
\newenvironment{tablehere} 
    {\def\@captype{table}} 
    {} 
\makeatother
\makeatletter
\def\@captype{table}
\makeatother

\begin{center}
\setlength{\tabcolsep}{0.5mm}
\centering 
\caption{
The Point group for each system.
}
\resizebox{0.30\textwidth}{!}{
\begin{tabular}{cccc}
\toprule
\textbf{System} & \textbf{Point group} & \textbf{\#Elements}& \textbf{\#Dimension} \\
\midrule
\textbf{Particle-Cube} & $O_h$ &48&3 \\
\textbf{Particle-Square prim} & $D_{4h}$ &16&3 \\
\textbf{Particle-Rectangle prism} & $D_{2h}$ &8&3 \\
\textbf{LiPS} & $C_{i}$ &2&3 \\
\textbf{Li\textsubscript{4}P\textsubscript{2}O\textsubscript{7}} & $D_{2h}$ &8 &3\\
\textbf{Crowd} & $D_{2}$ &4 &2\\
\textbf{Vehicle} &$D_{2}$ &4 &2\\
\bottomrule
\end{tabular}
}\label{table:group_appendix}
\vspace{-2ex}
\end{center}

\begin{center}
\setlength{\tabcolsep}{0.3mm}
\centering 
\caption{
Dataset statistics for Crowd and Vehicle dataset.
}
\resizebox{0.4\textwidth}{!}{
\begin{tabular}{cccccc}
\toprule
\textbf{Dataset} & \textbf{\#Sample} & \textbf{Mean \#Ped.} & \textbf{Max \#Ped.} & \textbf{Target frame}\\
\midrule
    \textbf{Ind.-Low} &1938 &23 &42 &10  \\
    \textbf{Ind.-High} &1961 &23 &37 &10  \\
    \textbf{Group-Low} &2121 &25 &44 &10  \\
    \textbf{Group-High} &2142 &24 &42 &10  \\
    \textbf{Highway 1} &2000 &10 &19 &100 \\
    \textbf{Highway 2} &2000 &9 &13 &100 \\
  \textbf{Highway 3} &2000 &13 &22 &100 \\
  \textbf{Highway 4} &2000 &31 &39 &100 \\
  \textbf{Highway 5} &2000 &13 &26 &100 \\
  \textbf{Highway 6} &2000 &20 &30 &100 \\
\bottomrule
\end{tabular}
}\label{table:dataset_appendix_2}
\end{center}
\begin{center}
\setlength{\tabcolsep}{0.30mm}
\centering 
\caption{
Dataset statistics for particle and molecular dataset.
}
\begin{adjustbox}{width=0.30\textwidth}
\begin{tabular}{cccccc}
\toprule
\textbf{Dataset} & \textbf{\#Sample} & \textbf{\#Atom} & \textbf{Target frame}\\
\midrule
\textbf{Particle}  &4200 &5 &10 \\
\textbf{LiPS}  &6000 &83 &10 \\
\textbf{Li\textsubscript{4}P\textsubscript{2}O\textsubscript{7}}  &6000 &208 &10 \\
\bottomrule
\end{tabular}
\end{adjustbox}
\label{table:dataset_appendix_1}
\end{center}

\begin{table*}[t]
\setlength{\tabcolsep}{1mm}
\centering 
\caption{
Ablation Studies on model designs. Bold font indicates the best result. We report both mean and standard deviation that are computed over 5 runs.
}
\vspace{-2ex}
\label{table:ablation_appendix}
\resizebox{0.95\textwidth}{!}{
\begin{tabular}{c|cccc|cccccc|cc}
\toprule
 \multirow{2}{*}{\textbf{Model}}  & \multicolumn{4}{c|}{\textbf{Crowd} ($\times 10^{-3}$)} & \multicolumn{6}{c|}{\textbf{Vehicle} ($\times 10^{-2}$)} & \multicolumn{2}{c}{\textbf{Molecular}($\times 10^{-1}$)}\\
 & Ind.-Low & Ind.-High & Group-Low & Group-High & Highway 1 & Highway 2 & Highway 3 & Highway 4 & Highway 5 & Highway 6 & LiPS & Li\textsubscript{4}P\textsubscript{2}O\textsubscript{7}\\
\midrule
  \textbf{E(n)}  & 0.65{$\pm$ 0.09} & 0.96{$\pm$0.08} & 1.09{$\pm$ 0.24} & 0.98{$\pm$ 0.09} & 0.05{$\pm$0.02} & 0.11{$\pm$0.05}& 0.14{$\pm$0.05} & 0.72{$\pm$0.11}  & 0.14{$\pm$0.03} & 0.54{$\pm$0.53} & 5.44{$\pm$ 0.33} & 12.1{$\pm$ 0.46}\\
  \textbf{O(n)} & 0.61{$\pm$ 0.05} & 0.92{$\pm$0.11} & 1.59{$\pm$ 0.21} & 1.00{$\pm$ 0.08} & \textbf{0.04{$\pm$0.04}} & 0.05{$\pm$0.02}& \textbf{0.13{$\pm$0.08}} & 0.80{$\pm$0.11}  & 0.16{$\pm$0.07} & 0.51{$\pm$0.53} & 6.46{$\pm$ 0.33} & 10.7{$\pm$ 0.89}\\
  \textbf{T(n)} & 0.83{$\pm$ 0.10} & 1.26{$\pm$0.08} & 1.79{$\pm$ 0.14} & 1.11{$\pm$ 0.06} & 0.13{$\pm$0.09} & 0.09{$\pm$0.04}& 0.25{$\pm$0.14} & 0.99{$\pm$0.14}  & 0.18{$\pm$0.05} & 0.43{$\pm$0.20} & 5.07{$\pm$ 0.08} & 10.8{$\pm$ 0.51}\\
  \midrule
    \textbf{Ranking} & 0.34{$\pm$ 0.02} & 0.50{$\pm$0.01} & 0.73{$\pm$ 0.09} & 0.37{$\pm$ 0.03} & 0.10{$\pm$0.04}& 0.06{$\pm$0.01}& 0.20{$\pm$0.02}& 1.11{$\pm$0.40}& 0.16{$\pm$0.01}& 0.61{$\pm$0.29} & 4.63{$\pm$ 1.19} & \textbf{9.18{$\pm$ 0.74}} \\
  \textbf{Sum Pooling} & 0.33{$\pm$ 0.02} & 0.42{$\pm$0.02} &\textbf{ 0.61{$\pm$ 0.09}} & 0.35{$\pm$ 0.03} & 1.95{$\pm$ 0.06} & 2.37{$\pm$ 0.97} & 3.26{$\pm$ 3.10} & 20.8{$\pm$ 15.1} & 5.46{$\pm$ 5.30} & 9.84{$\pm$ 2.56} & 4.37{$\pm$ 0.53} & 9.79{$\pm$ 0.97} \\
   \textbf{Mean Pooling} & 0.33{$\pm$ 0.05} & 0.52{$\pm$0.03} & 0.74{$\pm$ 0.13} &\textbf{ 0.34{$\pm$ 0.04}} & 0.19{$\pm$ 0.13} & 0.05{$\pm$ 0.05}& 0.53{$\pm$0.5} & 1.45{$\pm$ 0.59} & 0.14{$\pm$0.06} & 0.78{$\pm$ 0.53} & \textbf{4.11{$\pm$ 0.14}} & 9.80{$\pm$ 0.54}\\
\midrule
\textbf{DEGNN} & \textbf{0.32{$\pm$ }0.04} & \textbf{0.41{$\pm$0.04}} & 0.71{$\pm$0.13} & 0.36{$\pm$ 0.04} & 0.07{$\pm$ 0.04} & \textbf{0.04{$\pm$ 0.01}}  &  0.19{$\pm$ 0.05} & \textbf{0.56{$\pm$ 0.08}} & \textbf{0.11{$\pm$ 0.02}}  &  \textbf{0.27{$\pm$ 0.04}} & 4.36{$\pm$ 0.31} & 9.86{$\pm$ 0.97}\\
\bottomrule
\end{tabular}
}
\end{table*}

\begin{table*}[t]
\setlength{\tabcolsep}{1mm}
\centering 
\vspace{-1ex}
\caption{
Forward time in seconds for a batch size of 100 samples in LiPS dataset on GeForce RTX 4090 GPU.
}
\resizebox{0.8\textwidth}{!}{
\begin{tabular}{c|ccccccccccccc}
\toprule
\textbf{Model} & \textbf{GNN}& \textbf{S-LSTM}& \textbf{TFN}& \textbf{Radial Field} & \textbf{GNS}& \textbf{TransF} & \textbf{SE(3)-Trans.}  & \textbf{EGNN} & \textbf{GMN} & \textbf{CrowdSim}  & \textbf{SAKE} & \textbf{EqMotion}& \textbf{DEGNN}\\
\midrule
\textbf{Time (s)} &0.0019 &0.00047 &OOM &0.0023&0.0025 &0.0029 &OOM&0.0033&0.0039&0.0026&0.0063&0.1201&0.0088 \\
\bottomrule
\end{tabular}
}\label{table:time_appendix}
\end{table*}

\subsubsection{\textbf{LiPS}}
Lithium phosphorus sulfide (LiPS) is a crystalline superionic lithium conductor related to battery development. We adopt this dataset from ~\cite{batzner20223}. The LiPS data consists of 83 atoms. The datasets comprise 25,001 frames. We randomly select 2000/2000/2000 samples for the training, validation, and test sets respectively in the first 6000 frame.

\subsubsection{\textbf{Li\textsubscript{4}P\textsubscript{2}O\textsubscript{7}}}
The Li\textsubscript{4}P\textsubscript{2}O\textsubscript{7} were generated using an ab-initio meltquench MD simulation. We adopt this dataset from ~\cite{batzner20223}. A stoichiometric crystal of 208 atom interact with each other in a periodic box of 10.4 × 14.0 × 16. We adopt this dataset from ~\cite{batzner20223}. The datasets comprise 25,000 frames. We randomly select 2000/2000/2000 samples for the training, validation, and test sets respectively in the first 6000 frame.

\subsubsection{\textbf{Crowd}}
we classify the dataset into low-imbalanced individuals, low-imbalanced groups, high-imbalanced individuals, and high-imbalanced groups. Detailed information is as follows. The scenes are visualized in Figure~\ref{fig:crowd_img}.
\begin{itemize}
\item \textbf{Low imbalance individual flow}: bidirectional pedestrian movement with the presence of individuals. A total of 36 people walk from left to right, while 18 people walk from right to left.
\item \textbf{Highly imbalance Individual flow}: bidirectional pedestrian movement with the presence of individuals. A total of 45 people walk from left to right, while 9 people walk from right to left.
\item \textbf{Lowly imbalance Group flow}: bidirectional pedestrian movement with the presence of group. A total of 36 people walk from left to right, while 18 people walk from right to left.
\item \textbf{Highly imbalance Group flow}: bidirectional pedestrian movement with the presence of group. A total of 45 people walk from left to right, while 9 people walk from right to left.
\end{itemize}

\begin{table}[t]
\setlength{\tabcolsep}{0.5mm}
\centering 
\caption{
MAE metrics of two baselines and our proposed DEGNN on the charged dataset and vehicle dataset.
}
\vspace{-2ex}
\label{table:metric}
\resizebox{0.5\textwidth}{!}{
\begin{tabular}{c|cccc|cccccc}
\toprule
 \multirow{2}{*}{\textbf{Model}}  & \multicolumn{4}{c|}{\textbf{Charged Particle}} & \multicolumn{6}{c}{\textbf{Vehicle} ($\times 10^{-2}$)} \\
 & (5,5,5) & (5,4,4)	 & (5,4,3) & (4,4,4) & Highway 1 & Highway 2 & Highway 3 & Highway 4 & Highway 5 & Highway 6\\
\midrule
   \textbf{GNN}  &0.340 &	0.372  &0.374 &0.397  &0.053  &0.052  &0.067  &0.083  &0.058 &0.061  \\
      \textbf{GMN} &0.243 &0.290  &0.331 &0.310  &0.059  &0.061  &0.131  &0.129  &0.107 &0.113  \\

\midrule
\textbf{DEGNN} & \textbf{0.239} & \textbf{0.282} & \textbf{0.325}& \textbf{0.307}  & \textbf{0.019} & \textbf{0.015} & \textbf{0.031} & \textbf{0.047}& \textbf{0.025}  & \textbf{0.034}\\
\bottomrule
\end{tabular}
}
\end{table}

\subsubsection{\textbf{Highway}}
The scenes of Highways can be seen in Figure~\ref{fig:highway_img}

\begin{figure}[t]
    \centering
    \includegraphics[width=0.5\textwidth]{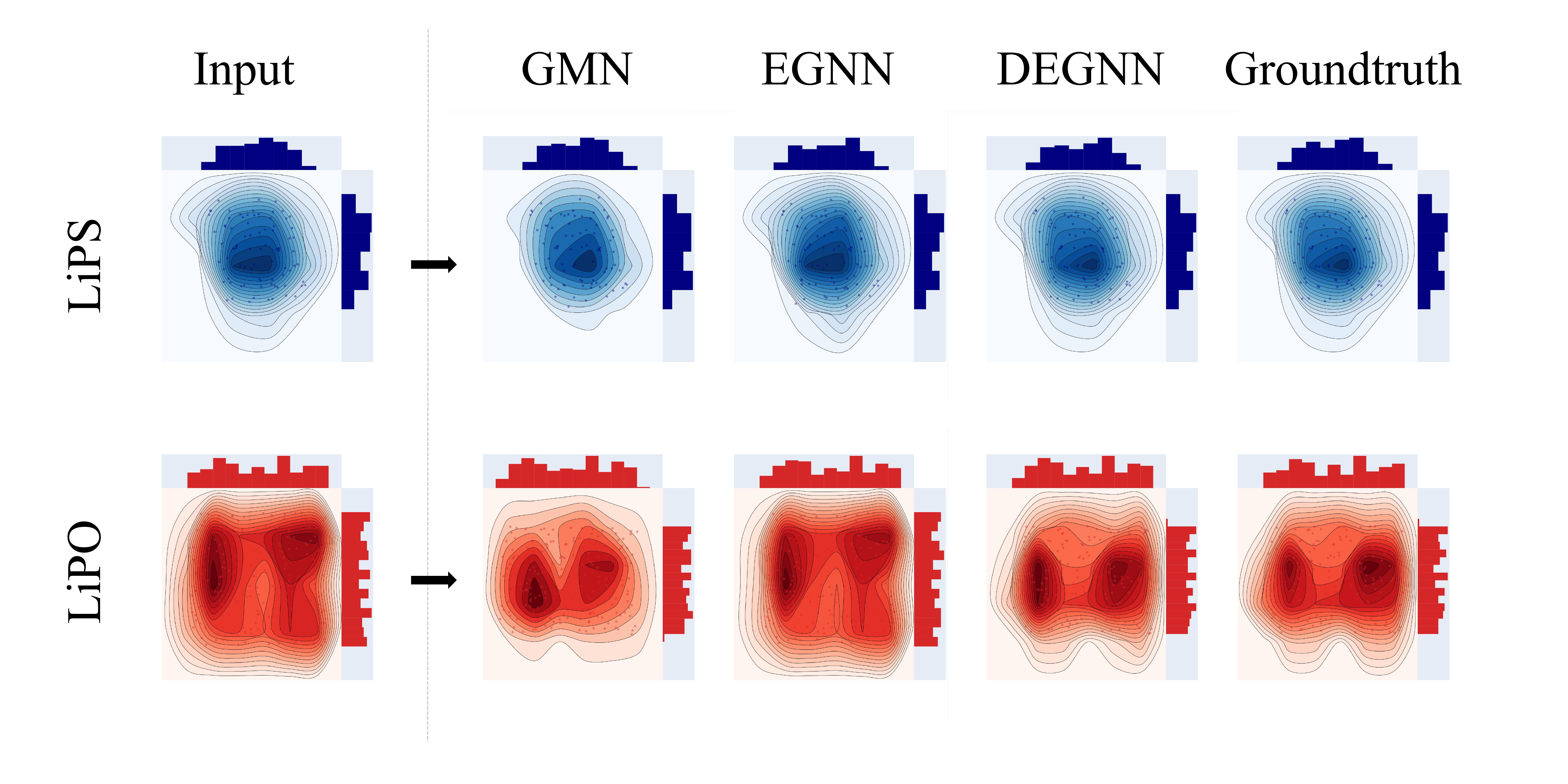}
    \vspace{-3ex}
    \caption{Visualization of DEGNN and other baselines prediction of molecular. Our method produces a molecular distribution that is closer to the groundtruth.}
    \label{fig:density}
\end{figure}

\section{Additional Results}

\subsection{Running Time Comparison}
To compare the efficiency, we report the running time of all methods in the LiPS dataset in Table~\ref{table:time_appendix}. The results show that DEGNN is faster than EqMotion and yields comparable results to SAKE. 

\subsection{Additional Ablation Study}
Table~\ref{table:ablation_appendix} shows the results of ablation studies on the other datasets. We can observe that the default settings of DEGNN outperform the other equivariant variants in almost all cases, verifying the idea of discrete equivariance. Additionally, compared with Ranking, Mean, and Sum Pooling variants, self-attention outperforms them in the vehicle and half of the crowd datasets, demonstrating its effectiveness on macro-level physical systems.

\subsection{Other evaluation metrics on model accuracy}
The MAE of two baselines and our proposed DEGNN across two datasets are shown in the Table~\ref{table:metric}.

\subsection{Visualization}
 To investigate deeper insight into DEGNN, we visualize predicted samples of molecular datasets by projecting the position onto the $xy$-plane. Figure~\ref{fig:density} shows their density. Based on the visualization results, we can find that DEGNN accurately recovers the distribution across all directions, whereas other models only capture patterns in specific directions. For example, in the LiPS dataset, the y-axis of EGNN output is similar to the ground truth, but it has an obvious difference on the x-axis. Such results again demonstrate the effectiveness of our model.

\end{document}